%% file: paper.tex
\documentclass[]{fairmeta}

\input{math_commands.tex}

\usepackage{hyperref}
\usepackage{url}
\usepackage{booktabs}   
\usepackage{multirow}   
\usepackage{adjustbox}
\usepackage{graphicx}
\usepackage{xcolor}
\usepackage{xspace}
\usepackage{pifont}
\usepackage{capt-of}
\usepackage{colortbl}
\usepackage{wrapfig}


\newcommand{\cmark}{\ding{51}}
\newcommand{\xmark}{\ding{55}}
\newcommand{\amark}{\textsc{auto}}  
\newcommand{\vepsilon}{\boldsymbol{\epsilon}}
\newlength{\benchnumw}
\newlength{\benchtail}
\newlength{\benchtailtrim}
\newcommand{\benchhead}[1]{\makebox[\benchnumw][l]{\hspace*{0.25\benchnumw}\makebox[0pt][l]{\rotatebox[origin=l]{50}{\scriptsize\textbf{#1}}}}}
\newcommand{\setbenchwidths}[1]{%
  \settowidth{\benchnumw}{\textbf{00.0}}%
  \settowidth{\benchtail}{\rotatebox[origin=l]{50}{\scriptsize\textbf{#1}}}%
  \addtolength{\benchtail}{-0.75\benchnumw}%
  \addtolength{\benchtail}{-\tabcolsep}%
  \ifdim\benchtail<0pt\setlength{\benchtail}{0pt}\fi
  \setlength{\benchtailtrim}{\benchtail}%
  \addtolength{\benchtailtrim}{\cmidrulekern}}

\newcommand{\modelname}{AVTR\xspace}
\newcommand{\imagtok}{\texttt{<imagine>}\xspace}
\newcommand{\imagendtok}{\texttt{</imagine>}\xspace}
\newcommand{\thinktok}{\texttt{<think>}\xspace}
\newcommand{\thinkendtok}{\texttt{</think>}\xspace}

\title{AVTR: Adaptive Multimodal Reasoning with Visual Imagination}

\author[1,2,*]{Xiaoqian Shen}
\author[1,\dagger]{Yunyang Xiong}
\author[1]{Changsheng Zhao}
\author[1]{Saksham Suri}
\author[1]{Chong Zhou}
\author[1]{Zechun Liu}
\author[1]{Zhipeng Cai}
\author[3]{Zhuang Liu}
\author[1]{Qi Qian}
\author[1]{Wei Wen}
\author[1]{Hongyu Xu}
\author[1]{Hanxian Huang}
\author[1]{Ernie Chang}
\author[1]{Yanbei Chen}
\author[1]{Vikas Chandra}
\author[1]{Raghuraman Krishnamoorthi}
\author[2,\dagger]{Mohamed Elhoseiny}
\author[1,\dagger]{Lemeng Wu}

\affiliation[1]{Meta AI}
\affiliation[2]{King Abdullah University of Science and Technology (KAUST)}
\affiliation[3]{Princeton University}

\contribution[*]{Work done at Meta}
\contribution[\dagger]{Project lead}

\abstract{%
Vision-Language Models increasingly rely on textual reasoning to solve advanced multimodal tasks. Despite its effectiveness, this paradigm faces two fundamental limitations. First, relying on textual reasoning alone can limit perceptual ability in vision-centric questions, where accurate answers depend on interpreting task-relevant visual evidence. Second, applying textual reasoning uniformly across inputs overlooks question-dependent variation in both reasoning modality and depth, often producing verbose and unnecessary traces when little logical inference is required. To this end, we propose \textbf{A}daptive \textbf{V}isual-\textbf{T}extual \textbf{R}easoning (\modelname), which introduces visual imagination as a complementary reasoning modality to textual thinking, enabling the model to revisit its visual observations and distill the salient evidence through generative visual reflection. Furthermore, we formulate reasoning-modality selection as a learned routing problem, allowing a single policy to adaptively harness the complementary strengths of visual reflection and textual reasoning in response to the perceptual and inferential demands of the input. Across 16 benchmarks, \modelname improves the average score by 4.5 points over Qwen3.5 baseline, while reducing inference latency by 33\%. It also scales effectively into a smaller size backbone, outperforming the evaluated 2B-scale baselines.
}

\date{\today}

\metadata[Correspondence]{\email{xiaoqian.shen@kaust.edu.sa},\email{yunyang@meta.com}}
\metadata[Project Page]{\url{https://xiaoqian-shen.github.io/AVTR/}}

\begin{document}

\maketitle

\section{Introduction}

Vision-Language Models (VLMs)~\citep{liu2023visual,zhu2024minigpt,wang2025internvl35,bai2025qwen25vl} are multimodal in what they observe but largely unimodal in how they reason. Although they perceive both visual and textual inputs, their reasoning process is predominantly carried out in autoregressive text tokens~\citep{wei2022chain,zhang2023multimodal,shao2024visual}. This paradigm is effective when relevant visual evidence has been captured, leaving only logical inference to be performed. However, in practice, the perceived visual input can be either incomplete or unfaithful, especially on challenging vision tasks. As an example, Qwen3.5-9B~\citep{qwen2026qwen359b} achieves only 34.7\% accuracy on MMSI-Bench~\citep{yang2026mmsi}, substantially below the 97\% human accuracy, while textual reasoning raises the score by only 0.3 points to 35.0\%. Such marginal gain from textual reasoning suggests that visual perception remains a key bottleneck, constraining what further reasoning can achieve with incomplete or unreliable visual evidence.

Several approaches have been explored in prior work to address the perceptual bottleneck. Explicit visual-tool methods zoom, crop, or augment the input, but require external execution and additional visual processing at inference~\citep{gupta2023visual,suris2023vipergpt,hu2024visual,yang2023set,shen2025zoom,zheng2026deepeyes,su2026pixel}. Unified models combine understanding and generation~\citep{dong2024dreamllm,zhou2025transfusion,xie2025showo}, but generation is rarely used to provide useful feedback for perception or reasoning, a direction explored only recently~\citep{li2025imagine,duan2026joyai}. Latent visual-reasoning methods extend reasoning beyond discrete text tokens to continuous, semantically aligned visual states~\citep{yang2026machine,li2026latent,pham2025multimodal,wang2026monet,shen2026visreflect}, yet a non-negligible gap remains between latent semantic alignment and faithful, question-conditioned visual perception.

In this work, we present \modelname, which introduces \textbf{visual imagination}
as an intermediate step to complement textual reasoning. It encourages the
model to revisit relevant visual evidence before generating a textual answer, rather
than reasoning directly from its initial visual observation. We enable this process
by bridging the output from a language model to the visual generative space, allowing it to
produce visual evidence, which is useful to the task. We further teach the model to distill
salient visual evidence through \textbf{generative visual reflection} and leverage
it for question-answering tasks. More importantly, the visual generator is solely used as the
training-time supervision and is completely removed at inference. Therefore, the model
is capable of reasoning from the learned visual states without decoding them
into pixels.

Furthermore, our empirical analysis reveals that visual imagination and textual
reasoning exhibit \textbf{complementary strengths} across different reasoning
demands (Figure~\ref{fig:route-preferences}). Visual imagination is particularly
effective for improving perception-intensive tasks by distilling task-relevant visual
evidence, whereas textual reasoning remains a great advantage for explicit relational
and compositional inference. This complementarity calls for an adaptive selection of
the reasoning route, in a way to best preserve answer correctness while avoiding
unnecessary computation. To this end, we propose \textbf{adaptive multimodal reasoning}, formulating it
as a learned routing problem over visual imagination and textual reasoning. A shared policy
selects direct answering, visual imagination, textual reasoning, or their
composition according to the demands of each question.

We perform extensive evaluations on various tasks, i.e. high-resolution image
understanding, spatial reasoning, video understanding, and temporal grounding over total 16 benchmarks. Across all the benchmarks, \modelname achieves strong average gain over different backbone modes, improving by a large margin of 4.9 points over direct answering and by 4.5 points over textual reasoning. In the meantime, we are able to reduce the average inference latency by 33\% w.r.t textual reasoning (Figure~\ref{fig:accuracy-latency}). Note that particularly on a 2B-parameter backbone, \modelname achieves a significant 6.3 points gain over the reference model. As a result, \modelname demonstrates that both visual
imagination and adaptive routing improve reasoning quality and
efficiency across different model scales.

\begin{figure*}[t]
\centering
\includegraphics[width=\textwidth]{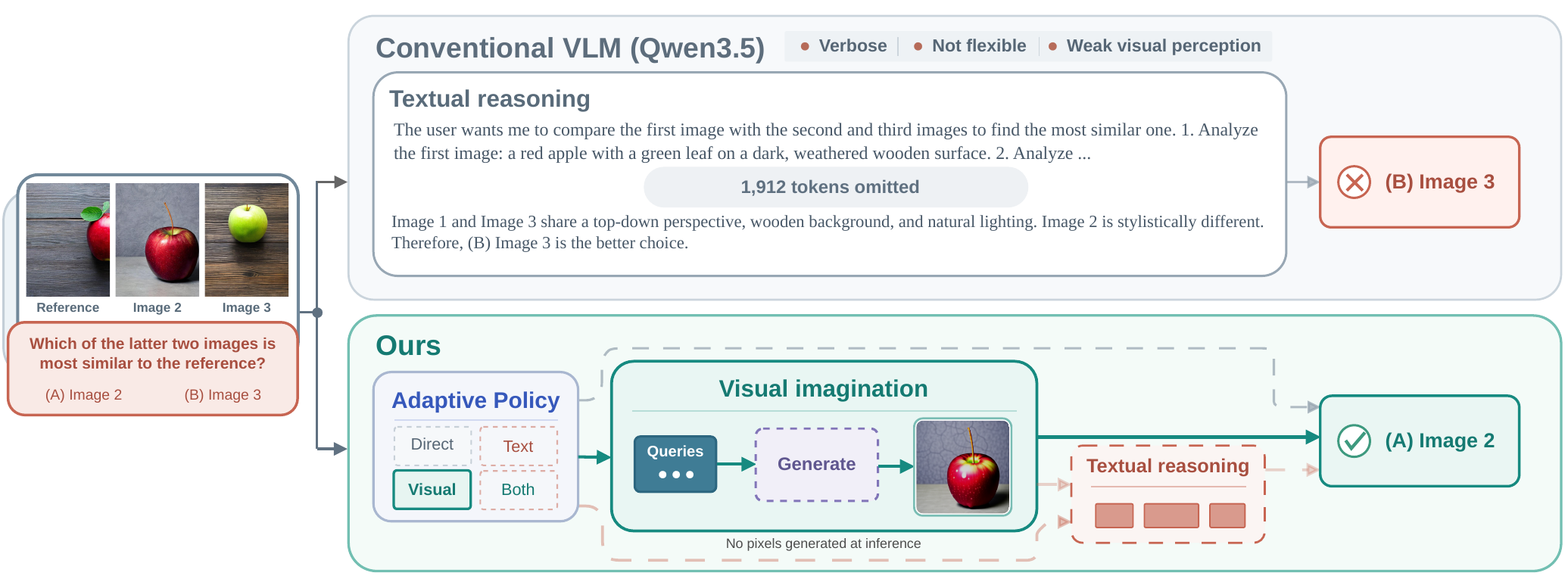}
\caption{\textbf{Adaptive multi-modal reasoning with visual imagination.}
Conventional VLMs perceive multimodal inputs but reason entirely in text.
\modelname{} adaptively harnesses the complementary strengths of visual imagination
and textual reasoning according to the perceptual and inferential demands of the
input. Visual imagination forms a question-conditioned visual state through
visual queries with training-time generative supervision, while requiring no
pixel generation at inference.}
\label{fig:teaser}
\end{figure*}

\noindent\textbf{Contributions.} Our contributions are summarized in threefold:

\begin{itemize}

\item We introduce visual imagination as an intermediate reasoning modality that revisits visual observations and distills task-relevant evidence for sophisticated vision tasks.

\item We formulate adaptive multi-modal reasoning as a learned routing problem over visual imagination and textual reasoning. Such design enables a single policy to select or compose the two based on each input's perceptual and inferential demands.

\item We demonstrate the effectiveness of \modelname on 16 benchmarks spanning diverse reasoning tasks. Our results show consistent improvements in
both evaluation metrics and inference efficiency. Meanwhile, it also scales effectively to a smaller 2B backbone.

\end{itemize}

\section{Related Work}

\noindent\textbf{Think with Vision.} VLMs combine visual encoders with language models and have achieved strong performance on image and video understanding \citep{liu2023visual,bai2025qwen25vl,li2024llavaonevision,tong2024cambrian}. Fine-grained visual questions remain challenging because attention can be drawn to visually salient but task-irrelevant tokens \citep{kang2025see}. Existing methods address this by explicitly selecting visual evidence, such as relevant regions or temporal spans, by using tools to crop, zoom, annotate, or re-encode selected regions \citep{gupta2023visual,suris2023vipergpt,hu2024visual,yang2023set,shen2025zoom,zheng2026deepeyes,su2026pixel}. While effective, these approaches typically introduce additional visual processing at inference. \modelname instead forms a question-conditioned visual state that revisits the original observation and distills task-relevant evidence.

\noindent\textbf{Latent Visual Reasoning.}
Latent reasoning replaces explicit textual traces with continuous states
\citep{hao2024training,goyal2024think}. Recent VLM methods align language-model
hidden states with visual features and feed them back as inputs for subsequent
reasoning
\citep{yang2026machine,li2026latent,pham2025multimodal,wang2026monet,shen2026visreflect,jeon2026vision}.
This output-to-input reuse may introduce a representation gap, while autoregressive
state generation adds sequential inference cost. In contrast, \modelname inserts
learnable visual queries directly into the VLM input and forms a
question-conditioned visual state in a single forward pass through generative
visual reflection in latent pixel.

\noindent\textbf{Adaptive Reasoning.}
Recent work has explored adapting textual reasoning to input difficulty. AdaptThink \citep{zhang2025adaptthink} learns to switch between Thinking and NoThinking modes, while VideoAuto-R1 \citep{liu2026videoauto} uses the confidence of an initial answer to decide whether further reasoning is needed. Other methods adapt textual computation through stopping or inference-time budgets \citep{chen2025not,sui2025stop,fang2025thinkless,muennighoff2025s1}. These approaches only focus on adapting
textual reasoning depth. In contrast, \modelname makes reasoning both multimodal and adaptive,
allowing the model to select between visual imagination and textual
reasoning based on route-level correctness and cost, while separately
optimizing reasoning within each route through route-conditioned GRPO.

\section{Method}
\label{sec:method}

\begin{figure*}[t]
\centering
\includegraphics[width=\textwidth]{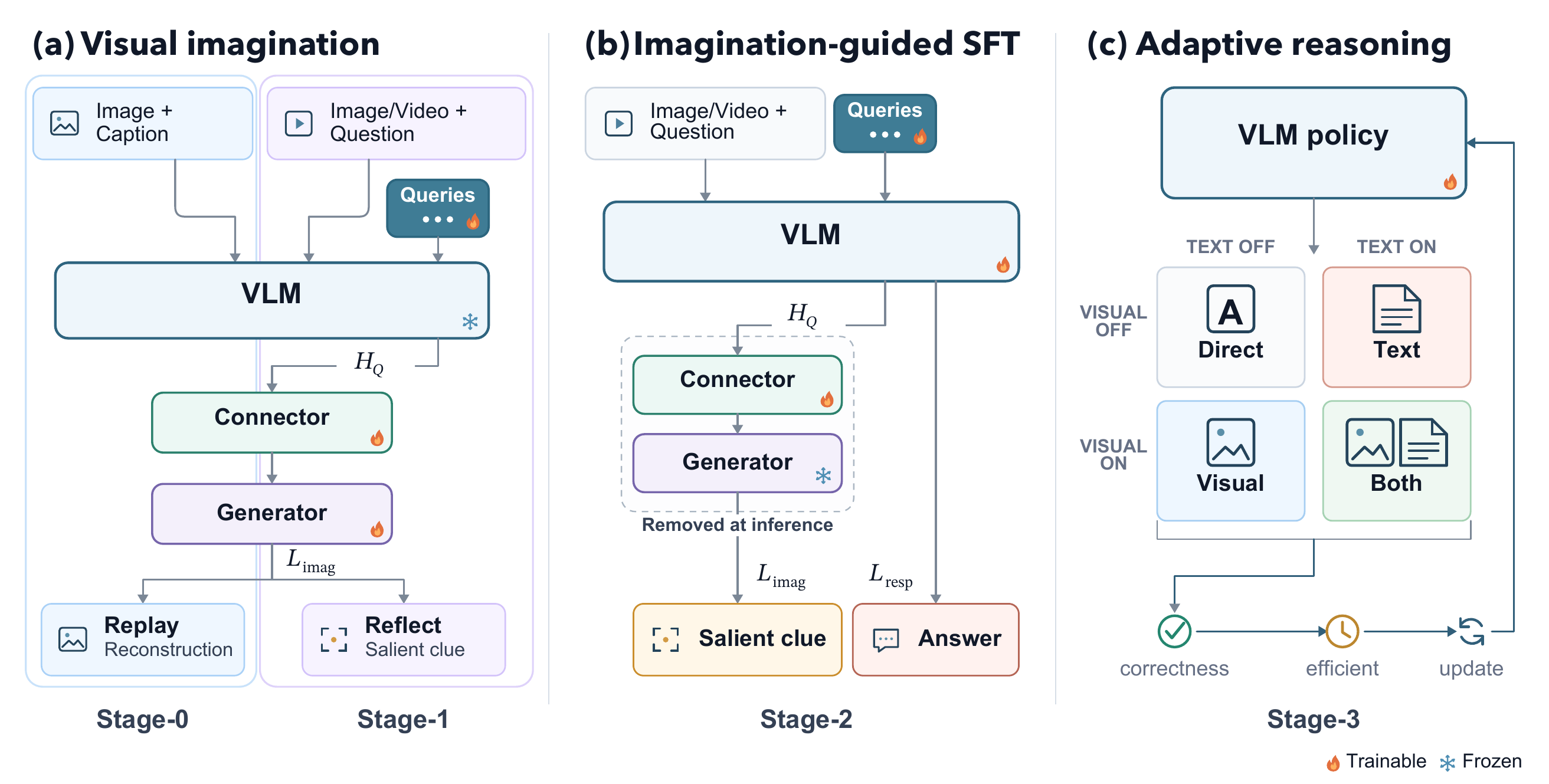}
\caption{\textbf{Method overview.}{
\textbf{(a) Visual imagination} is learned in two stages with shared trainable parameters but different inputs and supervision: \emph{visual replay} reconstructs the input from visual queries, while \emph{visual reflection} conditions them on the question to generate relevant visual evidence.
\textbf{(b) Imagination-guided SFT} jointly optimizes visual-generation and response objectives, enabling the resulting visual state to guide answer generation. The connector and generator provide training-only supervision and are removed at inference.
\textbf{(c) Adaptive reasoning} learns a correctness- and cost-aware policy that
selects direct answering, visual imagination, textual reasoning, or visual imagination
followed by textual reasoning for each input.}}
\label{fig:pipeline}
\end{figure*}

Given a visual input $\gV$ and a question $q$, a standard VLM $\pi_\theta$
predicts an answer $a$ from the visual and textual context. \modelname{} augments
this process with visual imagination, forming a question-conditioned visual state
that distills task-relevant visual evidence for answer generation. \textbf{Visual replay}
first aligns this state with the visual generative space, \textbf{visual reflection} makes
it question-dependent, and \textbf{imagination-guided} supervised finetuning teaches the VLM to leverage it
for answering. We then learn an outcome-guided \textbf{adaptive routing policy} from the empirical
performance of different reasoning routes. The policy adaptively selects among direct answering, visual imagination,
textual reasoning, and their composition for each input. Figure~\ref{fig:pipeline} gives an overview.

\subsection{Visual Imagination Through Replay and Reflection}
\label{sec:alignment}

\noindent\textbf{Visual Replay.}
As illustrated in Figure~\ref{fig:pipeline}(a), visual replay first learns a set of visual states that connect the VLM representation space
to a visual generator through reconstruction~\citep{pan2025transfer}. Let
$\mQ\in\mathbb{R}^{n\times d}$ denote $n$ learnable visual queries with hidden
dimension $d$. The text prompt $p$ is either the associated image caption or a general reconstruction instruction, ``Please reconstruct this
image''. The VLM processes $[\gV;p;\mQ]$ in a single forward pass, and
$\mH_Q\in\mathbb{R}^{n\times d}$ denotes the last hidden states at the
query positions. A connector $g_\psi$, parameterized by $\psi$, maps $\mH_Q$
to the conditioning space of the visual generator $v_\phi$, parameterized by
$\phi$. We optimize
\begin{equation}
\mathcal{L}_{\mathrm{imag}}
=
\mathbb{E}
\left[
\left\|
v_\phi\!\left(
\vz_\sigma,\sigma,g_\psi(\mH_Q)
\right)
-
(\vepsilon-\vz_0)
\right\|_2^2
\right],
\qquad
\vz_\sigma=(1-\sigma)\vz_0+\sigma\vepsilon,
\label{eq:imagination-loss}
\end{equation}
where $\vz_0$ is the latent encoding of the visual target produced by the
frozen autoencoder, $\vepsilon\sim\mathcal{N}(0,I)$ is Gaussian noise,
$\sigma\sim\mathcal{U}(0,1)$ is the flow interpolation step, and
$\vz_\sigma$ is the interpolated latent at step $\sigma$. During visual replay,
the visual target is the input $\gV$ itself.

We optimize the visual queries, connector, and generator while keeping the VLM
and autoencoder frozen. Since the queries are forward all at once, this stage
learns a visual state without autoregressive generation of latent tokens.
Reconstructing the visual input from $g_\psi(\mH_Q)$ encourages $\mH_Q$ to recover what the model observes as visual input.

\noindent\textbf{Visual Reflection.}
Visual reflection then reuses the same architecture but replaces generic
reconstruction with question-conditioned supervision, as shown in
Figure~\ref{fig:pipeline}(a). We use
the question as the prompt, $p=q$, and replace the reconstruction target with
a question-relevant visual target $\gV_{\mathrm{tgt}}$.

For images, $\gV_{\mathrm{tgt}}$ is the salient region or visual evidence of the input visual context; for videos, it is the frame containing the faithful evidence. Training with these
targets encourages $\mH_Q$ to focus on the visual evidence needed to answer the
question. Appendix~\ref{app:visual-targets} describes target localization,
verification, and gain-based filtering.

\subsection{Imagination-Guided SFT}
\label{sec:guided-sft}

Generative alignment makes the query state visually decodable and focuses it on
question-relevant evidence, but does not directly supervise its role in answer
generation. To bridge this gap, imagination-guided SFT jointly optimizes response
supervision and the visual generation objective from the trained generator, as
illustrated in Figure~\ref{fig:pipeline}(b). For samples with a visual target,
we use $y=[\imagtok;\mQ;\imagendtok;a]$, where the continuous queries $\mQ$
are placed before the answer and provide visual evidence for answer generation.
These two special tokens are added to the VLM's vocabulary. For those without a
target, we use $y=[\imagendtok;a]$ and do not require $\mQ$ or
$\mathcal{L}_{\mathrm{imag}}$. Let $\mathcal{T}$ denote the supervised discrete
token positions in $y$, excluding $\mQ$. We optimize
\begin{equation}
\mathcal{L}_{\mathrm{resp}}
=
-\frac{1}{|\mathcal{T}|}
\sum_{t\in\mathcal{T}}
\log \pi_\theta(y_t\mid \gV,q,y_{<t}),
\qquad
\mathcal{L}_{\mathrm{SFT}}
=
\mathcal{L}_{\mathrm{resp}}
+
\lambda_{\mathrm{imag}}\mathcal{L}_{\mathrm{imag}},
\label{eq:guided-sft}
\end{equation}
where $\lambda_{\mathrm{imag}}$ controls the weight of the visual generation
objective.

The generative loss anchors the query state to the visual target, while the
response loss trains the model to leverage this state for answer generation.
The aligned generator, connector, and autoencoder are used only for training-time
visual supervision and are removed at inference. The visual query states remain
in the VLM context and directly condition answer generation.

\subsection{Adaptive Multimodal Reasoning}
\label{sec:routing}

Visual imagination provides visual evidence, whereas textual reasoning supports explicit compositional inference. We let the model decide when to
invoke each through two token gates. The visual gate predicts
$\ell_v\in\{\imagtok,\imagendtok\}$, and the textual gate predicts
$\ell_t\in\{\thinktok,\thinkendtok\}$. An opening token activates the corresponding
branch, while the closing token skips it. Together, the two gates define the route
space
$\mathcal M=\{\textsc{Direct},\textsc{Visual},\textsc{Textual},
\textsc{Visual+Textual}\}$.
For \textsc{Visual+Textual}, visual imagination precedes textual reasoning so that
the textual trace can attend to the visual state.

\noindent\textbf{Outcome-Guided Gate Learning.}
For each input $x=(\gV,q)$, we force all four routes in $\mathcal M$ and sample
$g$ continuations under each route, yielding $G=4g$ rollouts in total. Let
$i\in\{1,\ldots,G\}$ index all rollouts, and let $m_i\in\mathcal M$ denote the
route under which rollout $i$ is sampled. We use
$r_i^{\mathrm{ans}}\in\{0,1\}$ to denote its answer reward. The mean reward of
route $m$ is
\begin{equation}
R_m
=
\frac{1}{g}
\sum_{i:\,m_i=m}
r_i^{\mathrm{ans}}.
\label{eq:route-reward}
\end{equation}
Rather than retaining only the best-performing route, we preserve the relative
outcomes of all four routes through a soft routing target:
\begin{equation}
p(m\mid x)
=
\operatorname{Softmax}_{m\in\mathcal M}
\left(
R_m-C(m)
\right),
\label{eq:route-target}
\end{equation}
where $C(m)$ denotes the route-level cost used in the routing target. We use
$C(\textsc{Direct})<C(\textsc{Visual})<C(\textsc{Textual})
<C(\textsc{Visual+Textual})$, with $C(\textsc{Direct})=0$.
Table~\ref{tab:route-cost} (Appendix) analyzes the sensitivity of routing to these costs.

Since the gate tokens are forced rather than sampled during route-complete
evaluation, we supervise the routing decisions directly:
\begin{equation}
\mathcal L_{\mathrm{gate}}
=
-\sum_{m\in\mathcal M}p(m\mid x)
\left[
\log\pi_\theta(\ell_v^m\mid x)
+
\log\pi_\theta(\ell_t^m\mid x,\ell_v^m)
\right],
\label{eq:gate-loss}
\end{equation}
where $(\ell_v^m,\ell_t^m)$ denotes the pair of gate labels corresponding to
route $m$. The textual gate is evaluated after executing the visual branch
specified by $\ell_v^m$.

\noindent\textbf{Route-Conditioned Content Optimization.}
We use the same $G$ rollouts to optimize the remaining reasoning and answer
tokens with GRPO~\citep{shao2024deepseekmath}. We keep the standard GRPO
objective and modify only how the group-relative advantage is computed. Rather
than centering rewards across all routes, we center each rollout against those
sampled under the same route:
\begin{equation}
A_i
=
r_i^{\mathrm{ans}}
-
\operatorname{mean}
\left(
\left\{
r_j^{\mathrm{ans}}
:\,
m_j=m_i
\right\}
\right).
\label{eq:content-advantage}
\end{equation}
This prevents differences in route utility from entering the content update,
leaving route selection to $\mathcal L_{\mathrm{gate}}$. We omit
standard-deviation normalization to avoid over-weighting low-variance groups
during policy optimization~\citep{liu2025understanding}. Tokens at the two
gating positions, together with the continuous visual queries, are masked from
the GRPO loss. We denote the resulting loss by
$\mathcal L_{\mathrm{GRPO}}$. Rollouts with $A_i=0$ are excluded from
$\mathcal L_{\mathrm{GRPO}}$, as they provide no relative learning signal
within the route. The final objective is
\begin{equation}
\mathcal L
=
\mathcal L_{\mathrm{GRPO}}
+
\lambda_{\mathrm{gate}}\mathcal L_{\mathrm{gate}},
\label{eq:objective}
\end{equation}
where $\lambda_{\mathrm{gate}}$ controls the gate objective. The gate learns
which route to select, while GRPO learns how to reason within that route.

\begin{table*}[htbp]
    \centering
    \setlength{\tabcolsep}{4pt}
    \renewcommand{\arraystretch}{1.3}
    \caption{\textbf{High-resolution image understanding and spatial reasoning.}}
    \label{tab:main_results}
    \resizebox{\linewidth}{!}{%
    \begin{tabular}{@{}lcccccccc@{}}
        \toprule
        & \multicolumn{4}{c}{\textbf{HighRes Image}}
        & \multicolumn{4}{c}{\textbf{Visual Spatial}} \\
        \cmidrule(lr){2-5} \cmidrule(lr){6-9}
        \textbf{Model}
        & \textbf{V$^*$}
        & \textbf{HR-Bench-4K}
        & \textbf{HR-Bench-8K}
        & \textbf{BLINK}
        & \textbf{ERQA}
        & \textbf{MMSI-Bench}
        & \textbf{VSI-Bench}
        & \textbf{RealWorldQA} \\
        \midrule
        \multicolumn{9}{@{}l}{\textit{Proprietary}} \\
        GPT-5
        & 72.8 & - & - & 71.0 & 65.7 & 41.9 & 55.0 & 76.1 \\
        Gemini 2.5 Pro
        & 83.8 & 87.3 & 85.4 & 73.5 & 55.7 & 38.0 & 53.6 & 68.7 \\
        GPT-5 mini
        & 78.6 & 78.6 & 74.4 & 56.7 & 54.0 & 34.1 & 48.7 & - \\
        \midrule
        \multicolumn{9}{@{}l}{\textit{Open foundation models}} \\
        Qwen3.5-9B (w/o Think)
        & 86.3 & 84.7 & 82.5 & 70.0 & 46.0 & 34.7 & 57.2 & 75.4 \\
        Qwen3.5-9B (w/ Think)
        & 88.4 & 88.7 & 85.1 & 72.4 & 52.3 & 35.0 & 54.8 & 80.5 \\
        InternVL3.5-8B
        & 70.2 & 69.7 & 70.1 & 59.5 & 41.0 & 28.7 & 56.3 & 67.5 \\
        Cambrian-S-7B
        & 70.6 & 64.0 & 53.7 & - & 39.0 & 27.1 & 67.5 & 64.8 \\
        LLaVA-OneVision-2-8B
        & 72.0 & 72.0 & 63.1 & 63.5 & 43.3 & 29.6 & 70.9 & 69.7 \\
        JoyAI-Image-Und-8B
        & 85.9 & 78.8 & 76.5 & 69.6 & 45.0 & 35.8 & 60.1 & 74.0 \\
        \rowcolor{cyan!10}
        \modelname-9B (Ours)
        & \textbf{89.5}
        & \textbf{89.4}
        & \textbf{85.1}
        & \textbf{75.2}
        & \textbf{54.2}
        & \textbf{39.8}
        & \textbf{71.0}
        & \textbf{81.5} \\
        \bottomrule
    \end{tabular}}
\end{table*}

\begin{table*}[htbp]
      \centering
      \setlength{\tabcolsep}{4pt}
      \renewcommand{\arraystretch}{1.3}
      \caption{\textbf{Video understanding and temporal grounding.}}
      \label{tab:video}
      \resizebox{\linewidth}{!}{%
      \begin{tabular}{@{}l@{}>{\hspace{12pt}}cccc@{}>{\hspace{12pt}}cccc@{}}
          \toprule
          & \multicolumn{4}{c}{\textbf{Video Understanding} (Acc)}
          & \multicolumn{4}{c}{\textbf{Temporal Grounding} (mIoU)} \\
          \cmidrule(lr){2-5} \cmidrule(lr){6-9}
          \textbf{Model}
          & \textbf{MVBench} & \textbf{Video-MME} & \textbf{LongVideoBench} & \textbf{MLVU}
          & \textbf{Charades-STA} & \textbf{QVHighlights} & \textbf{ActivityNet-Cap.} & \textbf{NExT-GQA} \\
          \midrule
          Qwen3.5-9B (w/o Think) & 69.1 & 72.4 & 67.8 & 79.1 & 61.8 & 67.2 & 49.8 & 38.4 \\
          Qwen3.5-9B (w/ Think)  & 71.8 & 75.4 & 69.1 & 80.0 & 55.2 & 63.5 & 42.1 & 34.4 \\
          InternVL3.5-8B         & 72.1 & 66.0 & 62.1 & 70.2 & 27.8 & 31.3 & 31.3 & 25.1 \\
          Video-R1               & 65.5 & 61.8 & -    & -    & -    & -    & -    & -    \\
          Time-R1                & 65.9 & 54.2 & 55.8 & -    & 58.8 & -    & 52.1 & -    \\
          VITAL                  & -    & 64.1 & -    & -    & 59.9 & -    & 49.8 & 43.0 \\
          VideoChat-R1.5         & 70.6 & 65.2 & 61.4 & 70.1 & 60.6 & -    & 35.3 & -    \\
          VideoAuto-R1-8B           & 72.0 & 71.7 & 67.4 & 78.1 & 63.7 & 61.6 & 51.9 & 44.2 \\
          LLaVA-OneVision-2-8B   & 66.2 & 71.9 & 68.0 & 76.6 & 53.5 & 63.3 & 50.5 & 35.6 \\
          \rowcolor{cyan!10}\modelname-9B (Ours)
          & \textbf{72.4}
          & \textbf{76.9}
          & \textbf{70.5}
          & \textbf{81.7}
          & \textbf{64.7}
          & \textbf{71.5}
          & \textbf{53.1}
          & \textbf{44.8} \\
          \bottomrule
      \end{tabular}}
\end{table*}

\section{Experiments}
\label{sec:exp}

\subsection{Experimental Setup}

\noindent\textbf{Evaluation Benchmarks and Metrics.}
We evaluate high-resolution image understanding on V$^*$~\citep{wu2024vstar},
HR-Bench-4K/8K~\citep{wang2025divide}, and BLINK~\citep{fu2024blink};
spatial reasoning on ERQA~\citep{erqa2025}, MMSI-Bench~\citep{yang2026mmsi},
VSI-Bench~\citep{yang2025thinking}, and RealWorldQA~\citep{xai2024realworldqa};
and video understanding on MVBench~\citep{li2024mvbench},
Video-MME~\citep{fu2025videomme}, LongVideoBench~\citep{wu2024longvideobench},
and MLVU~\citep{zhou2025mlvu}. These benchmarks are evaluated using their
official accuracy metrics. We further evaluate temporal grounding using mIoU on
Charades-STA~\citep{gao2017tall}, QVHighlights~\citep{lei2021detecting},
ActivityNet Captions~\citep{krishna2017dense}, and NExT-GQA~\citep{xiao2024nextgqa}.
Appendix~\ref{app:evaluation} provides the baseline sources and complete evaluation
protocol.

\noindent\textbf{Implementation Details.}
We use Qwen3.5-9B and Qwen3.5-2B
\citep{qwen2026qwen359b,qwen2026qwen352b} as the VLM backbones, with
$n=64$ learnable visual queries and a 6-layer connector to SANA-1.6B~\citep{xie2025sana} at $512\times512$ resolution.
Training proceeds in four stages: visual replay on 5M captioned images with the
VLM frozen for alignment; visual reflection on 178K question-conditioned targets; SFT on 380K instruction examples; and adaptive routing on 20K
questions with $g=4$ rollouts per route. We set
$\lambda_{\mathrm{imag}}=0.2$ and $\lambda_{\mathrm{gate}}=0.02$.
All models use the same prompts and greedy decoding, with up to 20K visual tokens,
256 video frames sampled at 2 fps, and 8,192 newly generated tokens.
Appendix~\ref{app:training} and~\ref{app:evaluation} provide the full training
and evaluation details.

\subsection{Main Results}
\label{sec:main}

\noindent\textbf{Image and spatial understanding.}
Table~\ref{tab:main_results} shows consistent gains across high-resolution
understanding and spatial reasoning. Relative to direct answering with the backbone (w/o Think),
\modelname improves all eight benchmarks and raises the average from 67.1 to 73.2,
a 6.1-point gain. The improvement is particularly pronounced on spatial reasoning
($+8.3$ points), with the largest gain on VSI-Bench ($+13.8$ points). This pattern
is consistent with the intended role of visual imagination: generating
task-relevant visual evidence that supports visual perception and spatial reasoning.

\noindent\textbf{Video understanding and temporal grounding.}
The same benefit extends from spatial to temporal evidence
(Table~\ref{tab:video}). \modelname improves over direct answering by 3.3 points
on average for video understanding and by 4.2 points for temporal grounding,
suggesting better localization of task-relevant temporal evidence. We also observe
that enabling textual reasoning improves the backbone on video understanding but
consistently degrades performance across all four temporal-grounding benchmarks,
further motivating adaptive reasoning. For a fair comparison with manageable inference cost, we limit thinking models to
a maximum of 8,192 generated tokens rather than adopting the backbone's official
224K long-video setting.

\noindent\textbf{Effectiveness on a smaller backbone.}
\label{sec:compact-backbone}

\begin{table*}[!htbp]
    \centering
    \caption{\textbf{Results with small VLMs.} \modelname{} uses Qwen3.5-2B
    as its backbone.}
    \label{tab:small}
    \small
    \setlength{\tabcolsep}{3.5pt}
    \renewcommand{\arraystretch}{1.22}
    \setbenchwidths{MLVU}
    \begin{adjustbox}{max width=\linewidth}
    \begin{tabular}{l >{\hspace{2pt}}cccc >{\hspace{6pt}}cccc >{\hspace{6pt}}ccc c<{\hspace{\benchtail}}}
        \toprule
        & \multicolumn{4}{c}{\textbf{HighRes Image}}
        & \multicolumn{4}{c}{\textbf{Visual Spatial}}
        & \multicolumn{4}{c}{\textbf{Video Understanding}\hspace*{\benchtail}} \\
        \cmidrule(lr){2-5} \cmidrule(lr){6-9} \cmidrule(l{\cmidrulekern}r{\benchtailtrim}){10-13}
        \textbf{Model}
        & \benchhead{V$^*$} & \benchhead{HR-Bench-4K} & \benchhead{HR-Bench-8K} & \benchhead{BLINK}
        & \benchhead{ERQA} & \benchhead{MMSI-Bench} & \benchhead{VSI-Bench} & \benchhead{RealWorldQA}
        & \benchhead{MVBench} & \benchhead{Video-MME} & \benchhead{LongVideoBench} & \benchhead{MLVU} \\
        \midrule
        Gemma 4 E2B
        & 55.0 & 56.5 & 47.8 & 45.9 & 36.3 & 27.3 & 21.4 & 49.7 & 40.6 & 56.2 & 46.4 & 68.8 \\
        MiniCPM-V 4.6 (1.3B)
        & 75.4 & 71.9 & 65.6 & 56.7 & 39.0 & 28.0 & 37.0 & 63.3 & 55.8 & 58.9 & 55.8 & 65.2 \\
        InternVL3.5-2B
        & 68.1 & 61.4 & 58.4 & 52.1 & 38.8 & 25.7 & 54.0 & 59.9 & \textbf{68.2} & 57.5 & 52.1 & 65.0 \\
        Qwen3.5-2B (w/o Think)
        & 82.7 & 77.1 & 74.0 & 48.3 & 34.0 & 22.2 & 48.0 & 67.7 & 58.6 & 60.9 & 51.9 & 69.3 \\
        Qwen3.5-2B (w/ Think)
        & 81.7 & 75.4 & 73.8 & 60.3 & 42.8 & 28.8 & 31.0 & \textbf{70.5} & 59.6 & 55.4 & 49.3 & 70.2 \\
        \rowcolor{cyan!10}\modelname-2B (Ours)
        & \textbf{86.4} & \textbf{77.6} & \textbf{75.8} & \textbf{64.4}
        & \textbf{48.0} & \textbf{32.7} & \textbf{60.8} & 69.0
        & 61.3 & \textbf{62.4} & \textbf{59.1} & \textbf{72.3} \\
        \bottomrule
    \end{tabular}
    \end{adjustbox}
\end{table*}

Table~\ref{tab:small} shows that \modelname scales well to the
2B backbone. Across all benchmarks, it improves the average score by 6.3
points over base model, with consistent gains across high-resolution image,
spatial, and video understanding. These results demonstrate that adaptive multimodal reasoning
remains effective across model scales.

\subsection{Visual and Textual Reasoning Are Complementary}
\label{sec:adaptive}

\begin{table*}[htbp]
    \centering
    \caption{\textbf{Comparison of reasoning modalities.} Under
    \textbf{Think}, \cmark{} means that a modality is always used, \xmark{} that
    it is disabled, and \amark{} that the model decides per question.}
    \label{tab:ablation_mode}

    \small
    \setlength{\tabcolsep}{3.5pt}
    \renewcommand{\arraystretch}{1.22}
    \setbenchwidths{LongVideoBench}

    \begin{adjustbox}{max width=\linewidth}
    \begin{tabular}{
        l
        >{\centering\arraybackslash}m{1cm}
        >{\centering\arraybackslash}m{1cm}
        >{\hspace{2pt}}cccc
        >{\hspace{6pt}}cccc
        >{\hspace{6pt}}cc
        c<{\hspace{\benchtail}}
    }
        \toprule
        & \multicolumn{2}{c}{\textbf{Think}}
        & \multicolumn{4}{c}{\textbf{HighRes Image}}
        & \multicolumn{4}{c}{\textbf{Visual Spatial}}
        & \multicolumn{3}{c}{\textbf{Video Understanding}} \\
        \cmidrule(lr){2-3}
        \cmidrule(lr){4-7}
        \cmidrule(lr){8-11}
        \cmidrule(lr){12-14}

        \textbf{Model}
        & \textbf{Visual}
        & \textbf{Text}
        & \benchhead{V$^*$}
        & \benchhead{HR-Bench-4K}
        & \benchhead{HR-Bench-8K}
        & \benchhead{BLINK}
        & \benchhead{ERQA}
        & \benchhead{MMSI-Bench}
        & \benchhead{VSI-Bench}
        & \benchhead{RealWorldQA}
        & \benchhead{MVBench}
        & \benchhead{Video-MME}
        & \benchhead{LongVideoBench} \\
        \midrule

        Qwen3.5
        & \xmark & \xmark
        & 86.3 & 84.7 & 82.5 & 70.0
        & 46.0 & 34.7 & 57.2 & 75.4
        & 69.1 & 72.4 & 67.8 \\

        Qwen3.5
        & \xmark & \cmark
        & 88.4 & 88.7 & 85.1 & 72.4
        & 52.3 & 35.0 & 54.8 & 80.5
        & 71.8 & 75.4 & 69.1 \\

        \midrule

        \modelname
        & \xmark & \xmark
        & 87.9 & 84.7 & 82.8 & 73.3
        & 47.0 & 37.1 & 65.8 & 75.8
        & 70.0 & 74.3 & 68.5 \\

        \modelname
        & \cmark & \xmark
        & 88.4 & 86.7 & 83.8 & 75.9
        & 48.7 & 37.4 & 69.9 & 76.3
        & 70.3 & 74.4 & 69.4 \\

        \modelname
        & \xmark & \cmark
        & 89.5 & 88.3 & 84.5 & 73.1
        & 48.2 & 36.4 & 63.2 & 77.1
        & 72.1 & 74.9 & 70.1 \\

        \modelname
        & \cmark & \cmark
        & 88.5 & 87.8 & 83.5 & 74.2
        & 50.7 & 37.7 & 63.5 & 77.4
        & 70.4 & 74.4 & 69.2 \\

        \rowcolor{cyan!10}
        \modelname{} (default)
        & \makebox[0.55cm][c]{\amark}
        & \makebox[0.55cm][c]{\amark}
        & 89.5 & 89.4 & 85.1 & 75.2
        & 54.2 & 39.8 & 71.0 & 81.5
        & 72.4 & 76.9 & 70.5 \\

        \bottomrule
    \end{tabular}
    \end{adjustbox}
\end{table*}

\noindent\textbf{Reasoning-mode ablation.}
Table~\ref{tab:ablation_mode} compares different fixed reasoning routes with adaptive routing. Enabling visual imagination consistently improves over the
direct route, and replacing the learned visual state at inference removes this gain
(Table~\ref{tab:visual-state}, Appendix).
However, no fixed reasoning strategy is uniformly optimal: always using visual
imagination, textual reasoning, or their composition is inferior to adaptive routing
on average. In
particular, simply enabling more reasoning does not guarantee better performance.
The benefit therefore comes from selecting the appropriate reasoning modality for
each question rather than applying a fixed reasoning pattern globally.
Table~\ref{tab:route-invocation} (Appendix) further analyzes activation rates of the
visual and textual gates under the learned adaptive policy for each benchmark.

\begin{figure}[!htbp]
    \centering
    \includegraphics[width=\textwidth]{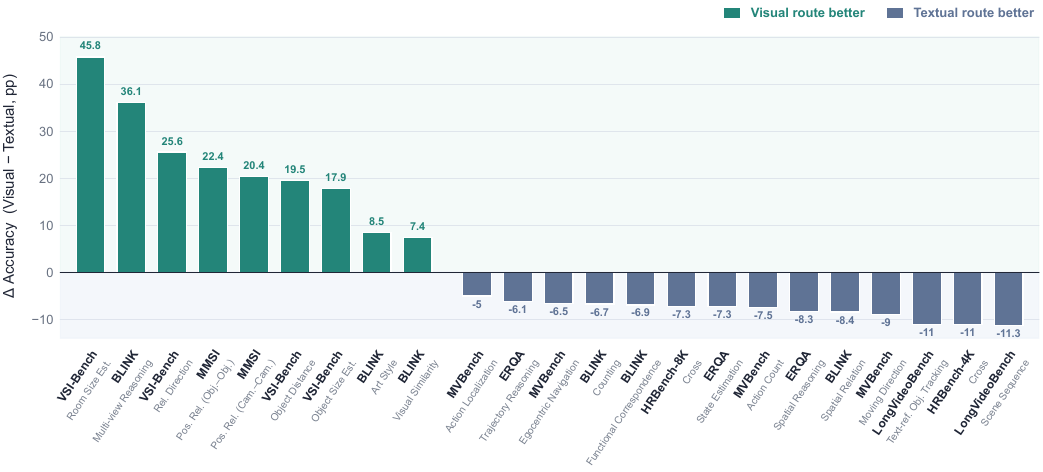}
    \caption{\textbf{Complementary strengths of visual and textual reasoning.} Accuracy differences between the visual imagination and textual reasoning on representative subtasks with large performance gaps. Positive values favor visual imagination, while negative values favor textual reasoning.}
    \label{fig:route-preferences}
\end{figure}

\noindent\textbf{Complementary strengths of visual and textual reasoning.}
Figure~\ref{fig:route-preferences} shows clear modality preferences across
representative subtasks. Visual imagination performs better on metric, geometric,
and viewpoint-sensitive perception, while textual reasoning is stronger on temporal
ordering, counting, and discrete relational reasoning. This complementarity motivates
adaptive routing to select the appropriate reasoning modality for each question.

\begin{figure}[!htbp]
    \centering
    \includegraphics[width=\textwidth]{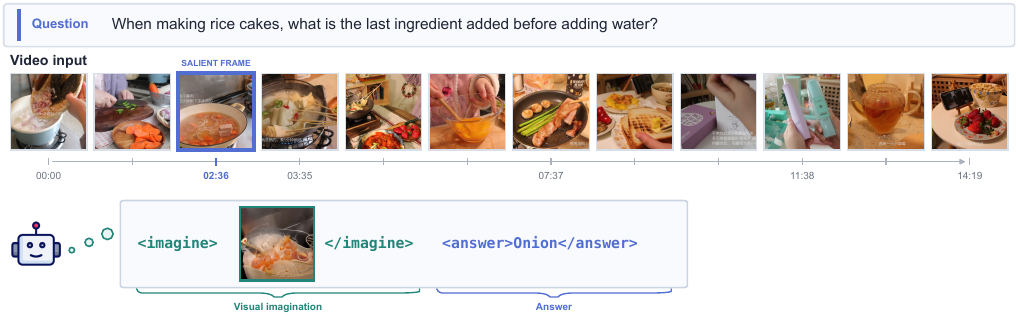}
    \caption{\textbf{Visual reflection for long-video understanding.} The rendered
    image is used only for visualization. The answer is conditioned directly on the
    continuous visual queries at inference.}
    \label{fig:video-qualitative}
\end{figure}

\begin{figure}[!ht]
    \makebox[\linewidth][l]{%
        \includegraphics[width=0.88\linewidth]{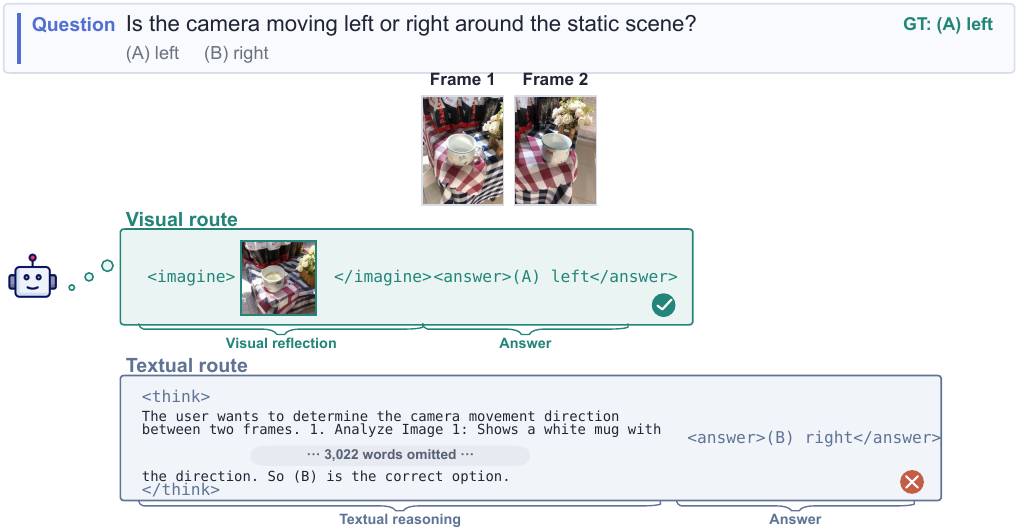}%
    }
    \caption{\textbf{Comparison between visual imagination and textual reasoning.}
    On this spatial task, textual reasoning produces a lengthy reasoning trace but
    still predicts the wrong camera-motion direction, whereas visual imagination
    reaches the correct answer with a compact visual state.}
    \label{fig:route-visual2}
\end{figure}

\noindent\textbf{Qualitative analysis.}
Figures~\ref{fig:video-qualitative} and~\ref{fig:route-visual2} illustrate
visual reflection for long-video understanding and the complementary strengths
of visual and textual reasoning on camera-motion questions. Additional temporal-grounding,
visual-similarity, and counting examples appear in Appendix~\ref{app:qualitative}
(Figures~\ref{fig:video-grounding-qualitative}, \ref{fig:route-visual}, and~\ref{fig:route-textual}).

\subsection{Ablation Studies}
\label{sec:training-ablation}

\begin{table*}[!ht]
    \centering
    \caption{\textbf{Ablation of visual supervision and model components.}
    The visual-supervision ablation uses the same instruction data without adaptive
    routing and changes only the visual supervision. The component ablation removes one
    component from the default model. Generative supervision and w/o adaptive routing
    are the same model.}
    \label{tab:training}
    \small
    \setlength{\tabcolsep}{6.08pt}
    \renewcommand{\arraystretch}{1.22}
    \setbenchwidths{LongVideoBench}
    \begin{adjustbox}{max width=\linewidth}
    \begin{tabular}{
        l
        >{\hspace{2pt}}cccc
        >{\hspace{6pt}}cccc
        >{\hspace{6pt}}cc
        c<{\hspace{\benchtail}}
    }
        \toprule
        & \multicolumn{4}{c}{\textbf{HighRes Image}}
        & \multicolumn{4}{c}{\textbf{Visual Spatial}}
        & \multicolumn{3}{c}{\textbf{Video Understanding}\hspace*{\benchtail}} \\
        \cmidrule(lr){2-5} \cmidrule(lr){6-9} \cmidrule(l{\cmidrulekern}r{\benchtailtrim}){10-12}
        \textbf{Variant}
        & \benchhead{V$^*$}
        & \benchhead{HR-Bench-4K}
        & \benchhead{HR-Bench-8K}
        & \benchhead{BLINK}
        & \benchhead{ERQA}
        & \benchhead{MMSI-Bench}
        & \benchhead{VSI-Bench}
        & \benchhead{RealWorldQA}
        & \benchhead{MVBench}
        & \benchhead{Video-MME}
        & \benchhead{LongVideoBench} \\
        \midrule
        Qwen3.5 (w/o Think)
        & 86.3 & 84.7 & 82.5 & 70.0 & 46.0 & 34.7 & 57.2 & 75.4 & 69.1 & 72.4 & 67.8 \\
        \midrule
        \multicolumn{12}{l}{\emph{Visual-supervision ablation}} \\
        No visual supervision
        & 86.9 & 82.3 & 80.6 & 72.1 & 47.7 & 37.5 & 65.3 & 75.0 & 69.5 & 72.1 & 68.7 \\
        Feature supervision
        & 87.4 & 82.3 & 79.1 & 72.5 & 48.0 & 37.2 & 65.1 & 75.6 & 69.8 & 72.9 & 69.4 \\
        Generative supervision
        & 88.4 & 86.7 & 83.8 & 75.9 & 50.7 & 38.4 & 67.9 & 76.3 & 71.5 & 74.4 & 69.4 \\
        \midrule
        \multicolumn{12}{l}{\emph{Component ablation}} \\
        w/o visual replay
        & 84.8 & 80.8 & 77.1 & 70.4 & 47.5 & 38.2 & 61.6 & 75.1 & 67.1 & 70.8 & 67.0 \\
        w/o visual reflection
        & 87.4 & 86.3 & 82.5 & 74.3 & 53.6 & 38.4 & 69.5 & 80.3 & 71.8 & 75.3 & 68.7 \\
        w/o adaptive routing
        & 88.4 & 86.7 & 83.8 & 75.9 & 50.7 & 38.4 & 67.9 & 76.3 & 71.5 & 74.4 & 69.4 \\
        \modelname{} (default)
        & 89.5 & 89.4 & 85.1 & 75.2 & 54.2 & 39.8 & 71.0 & 81.5 & 72.4 & 76.9 & 70.5 \\
        \bottomrule
    \end{tabular}
    \end{adjustbox}
\end{table*}

\noindent\textbf{Effect of generative visual supervision.}
The visual-supervision ablation in Table~\ref{tab:training} uses the same instruction
data without adaptive routing and changes only the visual supervision. No visual
supervision optimizes only $\mathcal{L}_{\mathrm{resp}}$. Feature supervision uses the
same curated visual targets and the same visual reflection and imagination-guided SFT
stages (Stage-1 and Stage-2) as our method, but replaces the generative objective with
feature-space cosine alignment to the visual target. Generative supervision
outperforms feature supervision on 10 of the 11 benchmarks and ties on the remaining
one, showing that generative grounding provides a stronger supervision signal for
learning task-relevant visual states.

\noindent\textbf{Contribution of each component.}
The component ablation removes one component from the default model: w/o visual
replay skips Stage-0, w/o visual reflection skips Stage-1, and w/o adaptive routing
skips Stage-3. Removing visual replay causes the largest degradation, suggesting that
question-conditioned reflection depends on a well-aligned interface between the VLM
and the visual generator. Visual reflection then makes the visual state
question-dependent, and adaptive routing learns when to invoke visual and textual
reasoning.

\subsection{Comparison With Prior Latent Visual Reasoning}
\label{sec:latent-comparison}

Figure~\ref{fig:latent-visual-comparison} (Appendix) compares both the effectiveness and
efficiency of visual imagination over latent visual reasoning methods. The reported $\Delta$ measures the
gain from enabling visual reasoning over the same method with visual reasoning disabled. Under this paired comparison, our generatively grounded visual imagination yields larger and
more consistent improvements across the four high-resolution benchmarks, with an
average gain of 1.53 points.
Our visual queries are inserted and processed in
parallel, avoiding token-by-token latent generation and reducing visual-reasoning
latency to 50\,ms versus 83-437\,ms for prior methods.

\subsection{Adaptive Reasoning Improves Performance and Efficiency}
\label{sec:exp-efficiency}

\begin{wrapfigure}{r}{0.37\textwidth}
\centering
\includegraphics[width=\linewidth]{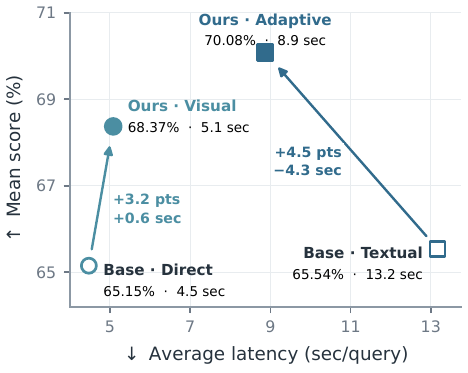}
\caption{\textbf{Efficiency analysis.} Mean score over the 16 benchmarks versus average
latency per query.}
\label{fig:accuracy-latency}
\end{wrapfigure}

Figure~\ref{fig:accuracy-latency} shows that visual imagination improves accuracy
with only 0.6 seconds of additional latency. This low overhead follows from two
design choices. (i) Visual queries are inserted and processed in parallel
rather than generated token by token. (ii) The generator is removed at inference,
so answers are conditioned directly on the intermediate visual states without pixel
decoding.

Adaptive reasoning further selects a route for each question and invokes textual
reasoning only when needed. In contrast, the backbone's fixed textual-reasoning mode
generates a full reasoning trace for every input. Combining visual and textual
reasoning on demand improves the mean score over the 16 benchmarks by 4.5 points
while reducing latency by 4.3 seconds. Appendix~\ref{app:efficiency} reports the
accuracy, latency, and number of generated tokens of each configuration.

\WFclear
\section{Conclusion}

We proposed \modelname, which introduces visual imagination as a complementary
reasoning modality and adaptively selects how to reason for each question.
Generative visual states revisit task-relevant evidence and directly
condition answer generation without pixel decoding at inference. A learned routing
policy selects or composes visual imagination and textual reasoning according to the
perceptual and inferential demands of each input. 
Experiments on various image and video understanding benchmarks
consistently validate the advantages of our model. We further show that the same framework remains effective with a
Qwen3.5-2B backbone, enabling a lightweight adaptive visual-textual reasoning model
and highlighting the broader applicability of \modelname across model scales.

\bibliographystyle{assets/plainnat}
\bibliography{paper}

\clearpage
\beginappendix

\setlength{\textfloatsep}{10pt plus 2pt minus 2pt}
\setlength{\floatsep}{8pt plus 2pt minus 2pt}
\setlength{\intextsep}{8pt plus 2pt minus 2pt}
\captionsetup{skip=4pt}

\section{Training Data and Optimization}
\label{app:training}

The training contains four stages: visual replay, question-conditioned reflection,
imagination-guided SFT, and adaptive routing. All phases use bfloat16 precision on
NVIDIA A100 80GB GPUs. We report the settings that determine optimization and
effective data scale, omitting infrastructure-specific parameters.

\subsection{Training Schedule}

\begin{figure*}[!htbp]
    \centering
    \includegraphics[width=\textwidth]{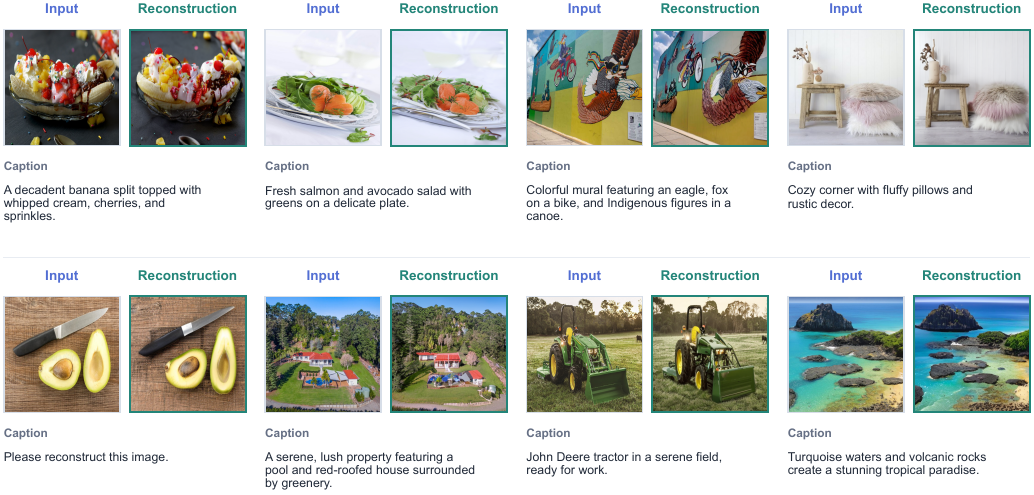}
    \caption{\textbf{Visual Reconstruction Quality.} Each example shows an input
    image, its associated image caption, and its reconstruction image.}
    \label{fig:stage0-reconstruction}
\end{figure*}

\paragraph{Stage-0: Visual Replay.}
Visual replay is trained on approximately 5M captioned images from
BLIP3o-Pretrain-Short-Caption~\citep{chen2025blip3o},
CC12M~\citep{changpinyo2021conceptual}, and
InternVL-SA-1B-Caption~\citep{kirillov2023segment,chen2024internvl,chen2024far}
on 128 A100 GPUs for 50K steps. The Qwen3.5-9B backbone and visual autoencoder
are frozen. We optimize the SANA-1.6B generator, 64 visual queries, and 6-layer
connector with AdamW~\citep{loshchilov2019decoupled}, a peak learning rate of
$3\times10^{-5}$, and cosine decay. All targets are reconstructed at
$512\times512$ resolution. Figures~\ref{fig:stage0-reconstruction}
and~\ref{fig:token-budget-reconstruction} show representative reconstructions,
while Table~\ref{tab:query-budget-ssim} quantifies the effect of the query budget.

\paragraph{Stage-1: Visual Reflection.}
Starting from visual replay, we train on approximately 178K question-conditioned
targets retained by the pipeline in Appendix~\ref{app:visual-targets}. The targets
comprise salient regions for images and salient frames for videos. We retain the
same frozen backbone and autoencoder and update only the generator, queries, and
connector. Training uses AdamW with learning rate $3\times10^{-5}$ for one epoch
with 64 A100 GPUs.

\paragraph{Stage-2: Imagination-Guided SFT.}
The SFT mixture contains approximately 380K examples from the sources summarized in
Table~\ref{tab:sft-data}. It combines spatial question answering, general video
question answering, and temporal grounding. Target-free examples remain in the
mixture and contribute only the response loss. We freeze the generator and visual
autoencoder and optimize the VLM, visual queries, and connector for one epoch. We
use AdamW with learning rate $10^{-6}$, weight decay $0.1$, a global batch size of
128, and cosine decay with a 3\% warmup with 64 A100 GPUs.

\begin{table}[!htbp]
    \centering
    \small
    \setlength{\tabcolsep}{2.5em}
    \renewcommand{\arraystretch}{1.15}
    \caption{\textbf{Training data.} The imagination-guided SFT mixture contains
    approximately 380K image, spatial, and video examples, including 178K data with
    verified visual target.}
    \label{tab:sft-data}
    \begin{adjustbox}{max width=\linewidth}
    \begin{tabular}{@{}lrl@{}}
        \toprule
        \textbf{Type} & \textbf{Size} & \textbf{Details} \\
        \midrule
        Image & 28.2K & Visual-CoT~\citep{shao2024visual} \\
        \midrule
        Spatial & 191.2K & JoyAI-Image-OpenSpatial~\citep{duan2026joyai} \\
        \midrule
        \multirow{3}{*}{Video} & \multirow{3}{*}{160.5K}
        & Open-o3-Video~\citep{tong2025openo3},
        VideoChat3-LV116K~\citep{videochat3team2026} \\
        & & ActivityNet Captions~\citep{krishna2017dense},
        Charades-STA~\citep{gao2017tall} \\
        & & QVHighlights~\citep{lei2021detecting},
        NExT-GQA~\citep{xiao2024nextgqa} \\
        \bottomrule
    \end{tabular}
    \end{adjustbox}
\end{table}

For datasets that also serve as evaluation benchmarks, we train only on their
official training splits and evaluate on held-out splits.

\paragraph{Stage-3: Adaptive Routing Data and Optimization.}
We construct the routing dataset from the SFT training pool. We first perform
route-complete rollouts for each question and discard samples for which either all
four routes are correct or all four are incorrect, as these cases provide no
discriminative supervision for route selection. Among questions solved by exactly
one route, we balance the four corresponding route categories so that no reasoning
mode dominates the routing data. This results in a training set of 20K questions.

During routing optimization, we force all four routes and sample four rollouts per
route for each question. We train with a training batch size of 8, a learning rate
of $4\times10^{-6}$, and GRPO clipping coefficients
$\epsilon_l=\epsilon_h=0.2$. We set the route costs to
$C(\textsc{Direct})=0$, $C(\textsc{Visual})=0.01$,
$C(\textsc{Textual})=0.04$, and
$C(\textsc{Visual+Textual})=0.05$.
We set $\lambda_{\mathrm{gate}}=0.02$ based on the approximate gradient
magnitudes of the two objectives, keeping the weighted gate gradient on the same
order of magnitude as the GRPO gradient.

\section{Evaluation Protocol}
\label{app:evaluation}

\paragraph{Baselines.}
The proprietary baselines are GPT-5~\citep{openai2025gpt5}, GPT-5
mini~\citep{openai2025gpt5mini}, and Gemini 2.5
Pro~\citep{comanici2025gemini25}. The open foundation models are Qwen3.5-9B and
Qwen3.5-2B~\citep{qwen2026qwen359b,qwen2026qwen352b}, InternVL3.5-8B and
InternVL3.5-2B~\citep{wang2025internvl35},
Cambrian-S-7B~\citep{yang2025cambrians},
LLaVA-OneVision-2-8B~\citep{an2026llavaonevision2},
JoyAI-Image-Und~\citep{duan2026joyai}, Gemma 4
E2B~\citep{gemmateam2026gemma4}, and
MiniCPM-V4.6~\citep{openbmb2026minicpmv46}. The video-specialized baselines are
Video-R1~\citep{feng2025video}, Time-R1~\citep{wang2026timer1},
VITAL~\citep{zhang2026vital}, VideoChat-R1.5~\citep{yan2026videochat}, and
VideoAuto-R1~\citep{liu2026videoauto}. In the visual-supervision
ablation, the variants with no visual supervision and with feature supervision are
trained with the same backbone and instruction data. The
latent-reasoning comparison includes LVR~\citep{li2026latent},
Monet~\citep{wang2026monet}, and VisReflect~\citep{shen2026visreflect}. The
Qwen3.5 rows marked w/o Think and w/ Think use direct answering and the model's
built-in textual-reasoning mode, respectively.

\paragraph{Evaluation.}
All models are evaluated using the same prompts and greedy decoding. Videos are
sampled at 2 fps, with at most 256 frames and 20K visual tokens. We set a maximum
generation budget of 8,192 new tokens to handle rare cases where textual reasoning
fails to terminate. If the model reaches this limit without producing a valid final
answer, we fall back to the prediction from the same backbone with thinking
disabled, rather than scoring an incomplete reasoning trace.

\section{Visual Target Construction and Filtering}
\label{app:visual-targets}

We construct the reflection targets $\gV_{\mathrm{tgt}}$ through candidate
localization, target verification, and gain-based filtering. The raw pool contains
approximately 0.92M image and video question-answer pairs from the full set of data
in Table~\ref{tab:sft-data}. We infer missing targets through model-assisted
localization when possible and then verify all targets, yielding approximately 534K
examples: 449K images and 84K salient video frames. The final gain-based filter
retains approximately 178K examples for reflection training, comprising
approximately 155K image targets and 23.5K video targets. These statistics describe
the dedicated question-conditioned reflection set.

\paragraph{Candidate Localization.}
For image examples with a ground-truth bounding box, we directly use it to crop the
image as the target. When a bounding box is unavailable,
Qwen3.5-27B~\citep{qwen2026qwen3527b} predicts the region that supports the
reference answer. For video examples with an annotated temporal span, we sample a
candidate frame from that span. When a temporal span is unavailable, Qwen3.5-27B
first localizes the relevant interval, from which we sample the candidate frame. The
resulting cropped image or salient frame defines $\gV_{\mathrm{tgt}}$.

\paragraph{Target Verification.}
We next ask Qwen3.5-27B to verify each proposed localization against the original
visual input, question, and reference answer. A target passes verification only when
the marked region or sampled frame contains visual evidence that directly supports
the reference answer. This stage removes empty, irrelevant, and incorrectly
localized targets before they are used for training.

\paragraph{Gain-Based Filtering.}
Verification establishes that a target is relevant, but not that it improves the
model's answer. We therefore evaluate the same base VLM twice. The base model
receives the original visual input and question, $[\gV;q]$, while the
target-conditioned run appends the verified target,
$[\gV;q;\gV_{\mathrm{tgt}}]$. For multiple-choice and exact-match questions, we
retain wrong-to-correct transitions. For temporal grounding, we require a positive
increase in temporal IoU. This paired criterion ensures that each retained target
contributes measurable evidence beyond the original input.

\section{Ablation Studies}

\paragraph{Adaptive routing behavior.}
Table~\ref{tab:route-invocation} summarizes the routing decisions made by
\modelname{} across benchmarks.

\begin{table}[!htbp]
    \centering
    \small
    \setlength{\tabcolsep}{1.5em}
    \renewcommand{\arraystretch}{1.08}
    \caption{\textbf{Gate activation under adaptive routing.} Percentage of test
    examples for which the learned policy activates the visual or textual reasoning
    gate. The two rates need not sum to 100\%, since both gates can be activated for
    the same example or both can remain inactive.}
    \label{tab:route-invocation}
    \begin{adjustbox}{max width=\linewidth}
    \begin{tabular}{@{}lcc@{}}
        \toprule
        \textbf{Benchmark}
        & \textbf{Visual Gate ON (\%)}
        & \textbf{Textual Gate ON (\%)} \\
        \midrule
        V$^*$          & 22.9  & 34.4 \\
        HR-Bench-4K    & 31.0  & 77.1 \\
        HR-Bench-8K    & 25.4  & 74.6 \\
        BLINK          & 52.0  & 19.8 \\
        \addlinespace[2pt]
        ERQA           & 42.2  & 46.0 \\
        MMSI-Bench     & 30.4  & 11.9 \\
        VSI-Bench      & 74.3  & 10.8 \\
        RealWorldQA    & 51.6  & 71.8 \\
        \addlinespace[2pt]
        MVBench        & 66.0  & 45.6 \\
        Video-MME      & 83.9  & 52.8 \\
        LongVideoBench & 66.4  & 89.7 \\
        MLVU           & 89.0  & 49.0 \\
        \bottomrule
    \end{tabular}
    \end{adjustbox}
\end{table}

\begin{figure*}[!htbp]
    \centering
    \includegraphics[width=0.7\textwidth]{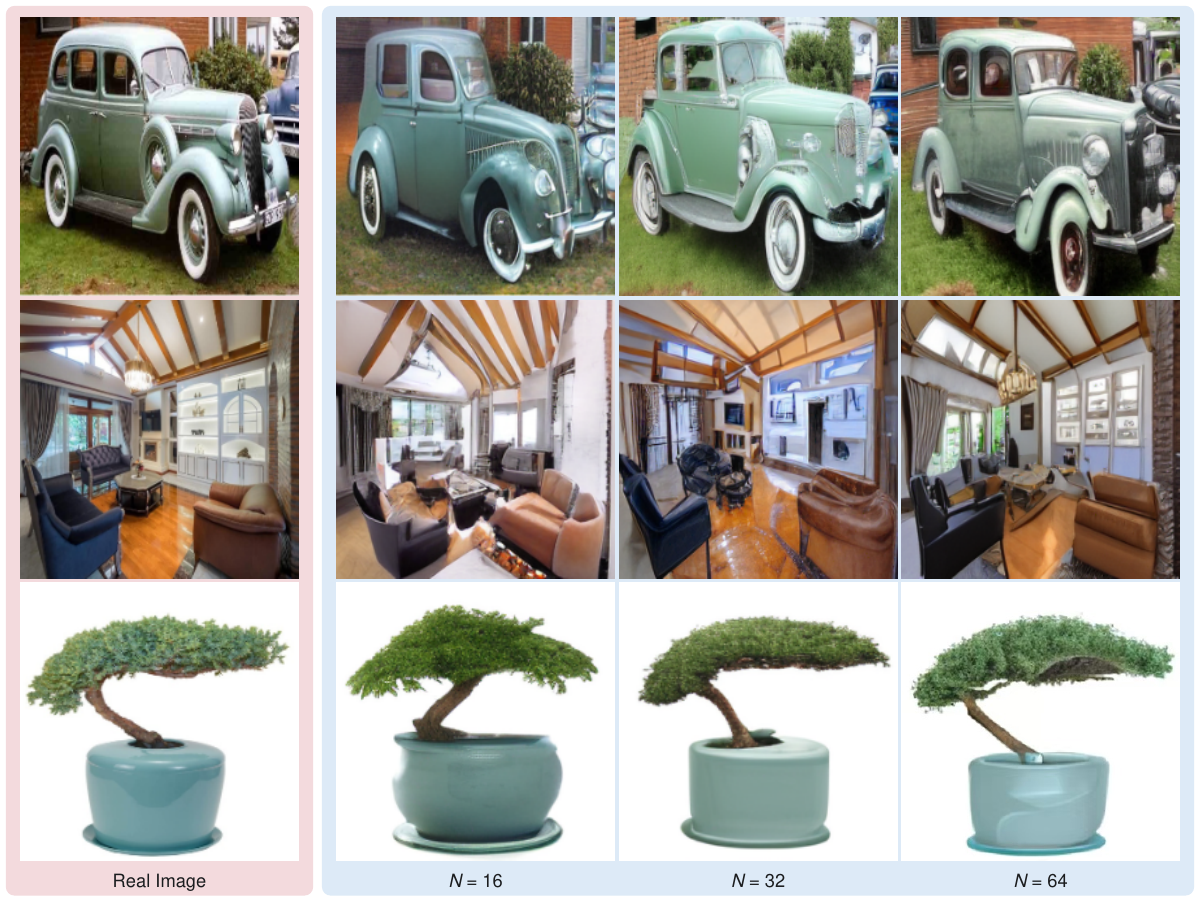}
    \caption{\textbf{Reconstruction quality across visual-query budgets.} Each row
    shows the input followed by reconstructions with $N=16$, $32$, and $64$ queries.}
    \label{fig:token-budget-reconstruction}
\end{figure*}

\begin{table}[!htbp]
    \centering
    \small
    \setlength{\tabcolsep}{1.4em}
    \renewcommand{\arraystretch}{1.15}
    \caption{\textbf{Effect of the visual-query budget on reconstruction.}
    Reconstruction fidelity measured by SSIM~\citep{wang2004image} improves
    consistently as the number of continuous visual queries increases.}
    \label{tab:query-budget-ssim}
    \begin{tabular}{@{}lccccc@{}}
        \toprule
        \textbf{Visual queries ($N$)} & 8 & 16 & 32 & 64 & 128 \\
        \midrule
        \textbf{SSIM} & 0.2563 & 0.3934 & 0.4217 & 0.5235 & 0.5632 \\
        \bottomrule
    \end{tabular}
\end{table}

\paragraph{Effect of the number of visual queries.}
Table~\ref{tab:query-budget-ssim} and
Figure~\ref{fig:token-budget-reconstruction} show that reconstruction quality
improves with the number of visual queries. SSIM increases from 0.2563 with 8
queries to 0.5235 with 64 queries and 0.5632 with 128 queries, while 64 queries
already produce visually faithful reconstructions. We therefore use 64 queries to
balance reconstruction quality and training cost, as larger query sets increase the
number of trainable $d$-dimensional query embeddings.

\paragraph{Sensitivity to the route cost.}
Table~\ref{tab:route-cost} varies the cost of the textual route,
$C(\textsc{Textual})$, around its default value of 0.04. $C(\textsc{Direct})$ and
$C(\textsc{Visual})$ stay at their default values, and $C(\textsc{Visual+Textual})$
changes together with $C(\textsc{Textual})$. Accuracy remains stable from 0.01 to 0.05, with averages over
the 12 benchmarks between 73.2 and 73.9, and decreases to 72.5 at 0.10. This is
consistent with the role of the route cost as a secondary efficiency preference, while
correctness remains the main routing signal.

\begin{table}[!htbp]
    \centering
    \caption{\textbf{Sensitivity to the route cost.} Accuracy of \modelname{} for different
    values of the textual-route cost $C(\textsc{Textual})$. $C(\textsc{Direct})$ and
    $C(\textsc{Visual})$ stay at their default values, and $C(\textsc{Visual+Textual})$
    changes together with $C(\textsc{Textual})$. Avg.\ is the mean over the 12 benchmarks.}
    \label{tab:route-cost}
    \small
    \setlength{\tabcolsep}{5.616pt}
    \renewcommand{\arraystretch}{1.1}
    \setbenchwidths{Avg.}
    \begin{adjustbox}{max width=\linewidth}
    \begin{tabular}{l >{\hspace{2pt}}cccc >{\hspace{6pt}}cccc >{\hspace{6pt}}cccc >{\hspace{6pt}}c<{\hspace{\benchtail}}}
        \toprule
        & \multicolumn{4}{c}{\textbf{HighRes Image}}
        & \multicolumn{4}{c}{\textbf{Visual Spatial}}
        & \multicolumn{4}{c}{\textbf{Video Understanding}}
        & \\
        \cmidrule(lr){2-5} \cmidrule(lr){6-9} \cmidrule(lr){10-13}
        $C(\textsc{Textual})$
        & \benchhead{V$^*$}
        & \benchhead{HR-Bench-4K}
        & \benchhead{HR-Bench-8K}
        & \benchhead{BLINK}
        & \benchhead{ERQA}
        & \benchhead{MMSI-Bench}
        & \benchhead{VSI-Bench}
        & \benchhead{RealWorldQA}
        & \benchhead{MVBench}
        & \benchhead{Video-MME}
        & \benchhead{LongVideoBench}
        & \benchhead{MLVU}
        & \benchhead{Avg.} \\
        \midrule
        0.01
        & 89.0 & 87.4 & 84.0 & 74.8 & 53.2 & 38.8 & 70.7 & 81.4 & 71.9 & 75.8 & 70.7 & 80.5 & 73.2 \\
        0.02
        & 90.1 & 89.4 & 84.9 & 74.0 & 53.6 & 37.5 & 70.3 & 81.5 & 71.4 & 75.0 & 70.1 & 81.0 & 73.2 \\
        \rowcolor{cyan!10}0.04 (default)
        & 89.5 & 89.4 & 85.1 & 75.2 & 54.2 & 39.8 & 71.0 & 81.5 & 72.4 & 76.9 & 70.5 & 81.7 & 73.9 \\
        0.05
        & 89.0 & 88.8 & 84.9 & 74.9 & 53.6 & 39.7 & 70.5 & 79.9 & 70.8 & 76.2 & 70.8 & 80.9 & 73.3 \\
        0.10
        & 89.0 & 87.1 & 83.5 & 71.4 & 54.7 & 38.9 & 70.3 & 78.3 & 70.6 & 75.8 & 69.8 & 80.7 & 72.5 \\
        \bottomrule
    \end{tabular}
    \end{adjustbox}
\end{table}

\paragraph{Intervening on the visual state.}
To test whether answers rely on the information in the visual state, we modify the
visual state at inference (Table~\ref{tab:visual-state}). Compared with the direct
route, which uses no visual state, the visual route improves accuracy on all 11
benchmarks. Replacing the learned visual queries with randomly initialized queries
lowers accuracy on every benchmark, to an average of 67.9, below the direct route
(69.7). Replacing them with the encoding of a randomly sampled image from the input
leads to a similar drop (68.7). The gain of the visual route therefore comes from
the task-relevant information carried by the learned visual state.

\begin{table}[!htbp]
    \centering
    \caption{\textbf{Intervening on the visual state at inference.} The direct route uses no
    visual state, and the visual route uses the learned visual queries. Random init queries
    replace the learned visual queries with randomly initialized queries, and encoded random
    image replaces them with the encoding of an image randomly sampled from the input.
    Avg.\ is the mean over the 11 benchmarks.}
    \label{tab:visual-state}
    \small
    \setlength{\tabcolsep}{5.19pt}
    \renewcommand{\arraystretch}{1.1}
    \setbenchwidths{Avg.}
    \begin{adjustbox}{max width=\linewidth}
    \begin{tabular}{l >{\hspace{2pt}}cccc >{\hspace{6pt}}cccc >{\hspace{6pt}}ccc >{\hspace{6pt}}c<{\hspace{\benchtail}}}
        \toprule
        & \multicolumn{4}{c}{\textbf{HighRes Image}}
        & \multicolumn{4}{c}{\textbf{Visual Spatial}}
        & \multicolumn{3}{c}{\textbf{Video Understanding}}
        & \\
        \cmidrule(lr){2-5} \cmidrule(lr){6-9} \cmidrule(lr){10-12}
        \textbf{Method}
        & \benchhead{V$^*$}
        & \benchhead{HR-Bench-4K}
        & \benchhead{HR-Bench-8K}
        & \benchhead{BLINK}
        & \benchhead{ERQA}
        & \benchhead{MMSI-Bench}
        & \benchhead{VSI-Bench}
        & \benchhead{RealWorldQA}
        & \benchhead{MVBench}
        & \benchhead{Video-MME}
        & \benchhead{LongVideoBench}
        & \benchhead{Avg.} \\
        \midrule
        Direct route
        & 87.9 & 84.7 & 82.8 & 73.3 & 47.0 & 37.1 & 65.8 & 75.8 & 70.0 & 74.3 & 68.5 & 69.7 \\
        Visual route
        & 88.4 & 86.7 & 83.8 & 75.9 & 48.7 & 37.4 & 69.9 & 76.3 & 70.3 & 74.4 & 69.4 & 71.0 \\
        \midrule
        Random init queries
        & 84.8 & 80.9 & 77.5 & 73.3 & 46.2 & 36.4 & 65.3 & 73.3 & 69.0 & 72.9 & 67.8 & 67.9 \\
        Encoded random image
        & 85.9 & 83.1 & 80.1 & 72.5 & 47.2 & 35.3 & 67.4 & 75.3 & 69.4 & 73.3 & 66.7 & 68.7 \\
        \bottomrule
    \end{tabular}
    \end{adjustbox}
\end{table}

\section{Inference Cost and Reasoning Length}
\label{app:efficiency}

\paragraph{Latency of each configuration.}
Table~\ref{tab:route-latency} reports the accuracy and the latency of each configuration
in Table~\ref{tab:ablation_mode}. At closely matched latency, adaptive
routing reaches an average accuracy of 73.2 at 8.9\,s, compared with 70.7 at 9.3\,s and
9.4\,s when always using the \textsc{Textual} and \textsc{Visual+Textual} routes,
respectively. Learned per-question selection therefore improves accuracy by about 2.5
points over the fixed routes of similar latency. It also outperforms the backbone with
textual reasoning by 2.9 points while reducing latency from 13.2\,s to 8.9\,s.

\begin{table}[!htbp]
    \centering
    \caption{\textbf{Accuracy and latency of each configuration.} The configurations and the
    per-benchmark accuracy are those of Table~\ref{tab:ablation_mode}.
    Avg.\ is the mean accuracy over the 11 benchmarks, and latency is the average inference
    time per query.}
    \label{tab:route-latency}
    \small
    \setlength{\tabcolsep}{2.408pt}
    \renewcommand{\arraystretch}{1.1}
    \setbenchwidths{Latency (s)}
    \begin{adjustbox}{max width=\linewidth}
    \begin{tabular}{
        l
        >{\centering\arraybackslash}m{1cm}
        >{\centering\arraybackslash}m{1cm}
        >{\hspace{2pt}}cccc
        >{\hspace{6pt}}cccc
        >{\hspace{6pt}}ccc
        >{\hspace{6pt}}c
        >{\hspace{4pt}}c<{\hspace{\benchtail}}
    }
        \toprule
        & \multicolumn{2}{c}{\textbf{Think}}
        & \multicolumn{4}{c}{\textbf{HighRes Image}}
        & \multicolumn{4}{c}{\textbf{Visual Spatial}}
        & \multicolumn{3}{c}{\makebox[0pt]{\textbf{Video Understanding}}}
        & & \\
        \cmidrule(lr){2-3}
        \cmidrule(lr){4-7}
        \cmidrule(lr){8-11}
        \cmidrule(lr){12-14}
        \textbf{Model}
        & \textbf{Visual}
        & \textbf{Text}
        & \benchhead{V$^*$}
        & \benchhead{HR-Bench-4K}
        & \benchhead{HR-Bench-8K}
        & \benchhead{BLINK}
        & \benchhead{ERQA}
        & \benchhead{MMSI-Bench}
        & \benchhead{VSI-Bench}
        & \benchhead{RealWorldQA}
        & \benchhead{MVBench}
        & \benchhead{Video-MME}
        & \benchhead{LongVideoBench}
        & \benchhead{Avg.}
        & \benchhead{Latency (s)} \\
        \midrule
        Qwen3.5
        & \xmark & \xmark
        & 86.3 & 84.7 & 82.5 & 70.0 & 46.0 & 34.7 & 57.2 & 75.4 & 69.1 & 72.4 & 67.8
        & 67.8 & 4.5 \\
        Qwen3.5
        & \xmark & \cmark
        & 88.4 & 88.7 & 85.1 & 72.4 & 52.3 & 35.0 & 54.8 & 80.5 & 71.8 & 75.4 & 69.1
        & 70.3 & 13.2 \\
        \midrule
        \modelname{}
        & \xmark & \xmark
        & 87.9 & 84.7 & 82.8 & 73.3 & 47.0 & 37.1 & 65.8 & 75.8 & 70.0 & 74.3 & 68.5
        & 69.7 & 4.3 \\
        \modelname{}
        & \cmark & \xmark
        & 88.4 & 86.7 & 83.8 & 75.9 & 48.7 & 37.4 & 69.9 & 76.3 & 70.3 & 74.4 & 69.4
        & 71.0 & 5.1 \\
        \modelname{}
        & \xmark & \cmark
        & 89.5 & 88.3 & 84.5 & 73.1 & 48.2 & 36.4 & 63.2 & 77.1 & 72.1 & 74.9 & 70.1
        & 70.7 & 9.3 \\
        \modelname{}
        & \cmark & \cmark
        & 88.5 & 87.8 & 83.5 & 74.2 & 50.7 & 37.7 & 63.5 & 77.4 & 70.4 & 74.4 & 69.2
        & 70.7 & 9.4 \\
        \rowcolor{cyan!10}\modelname{} (default)
        & \amark & \amark
        & 89.5 & 89.4 & 85.1 & 75.2 & 54.2 & 39.8 & 71.0 & 81.5 & 72.4 & 76.9 & 70.5
        & 73.2 & 8.9 \\
        \bottomrule
    \end{tabular}
    \end{adjustbox}
\end{table}

\paragraph{Number of generated tokens.}
Table~\ref{tab:route-tokens} reports the accuracy and the number of generated tokens for
the configurations with textual reasoning. Post-training substantially shortens the
textual reasoning traces: averaged over all questions of the 11 benchmarks, the number
of generated tokens decreases from 3,050 for the backbone with textual reasoning to 1,191
for the \textsc{Textual} route of \modelname{}, at a similar average accuracy (70.3 and
70.7). Adaptive routing generates 516 tokens on average, since it invokes textual
reasoning only on part of the questions (Table~\ref{tab:route-invocation}), while
achieving the highest average accuracy (73.2).

\begin{table}[!htbp]
    \centering
    \caption{\textbf{Accuracy and number of generated tokens.} For each configuration with
    textual reasoning, Acc.\ is the accuracy and Tokens is the average number of generated
    tokens per question. \textbf{Think} follows Table~\ref{tab:ablation_mode}. Avg.\ is the mean accuracy over the 11 benchmarks for Acc., and the average over
    all questions of the 11 benchmarks for Tokens.}
    \label{tab:route-tokens}
    \small
    \setlength{\tabcolsep}{2.514pt}
    \renewcommand{\arraystretch}{1.1}
    \setbenchwidths{Avg.}
    \begin{adjustbox}{max width=\linewidth}
    \begin{tabular}{
        l
        >{\centering\arraybackslash}m{1cm}
        >{\centering\arraybackslash}m{1cm}
        l
        >{\hspace{2pt}}cccc
        >{\hspace{6pt}}cccc
        >{\hspace{6pt}}ccc
        >{\hspace{6pt}}c<{\hspace{\benchtail}}
    }
        \toprule
        & \multicolumn{2}{c}{\textbf{Think}} &
        & \multicolumn{4}{c}{\textbf{HighRes Image}}
        & \multicolumn{4}{c}{\textbf{Visual Spatial}}
        & \multicolumn{3}{c}{\makebox[0pt]{\textbf{Video Understanding}}}
        & \\
        \cmidrule(lr){2-3}
        \cmidrule(lr){5-8}
        \cmidrule(lr){9-12}
        \cmidrule(lr){13-15}
        \textbf{Model}
        & \textbf{Visual}
        & \textbf{Text}
        &
        & \benchhead{V$^*$}
        & \benchhead{HR-Bench-4K}
        & \benchhead{HR-Bench-8K}
        & \benchhead{BLINK}
        & \benchhead{ERQA}
        & \benchhead{MMSI-Bench}
        & \benchhead{VSI-Bench}
        & \benchhead{RealWorldQA}
        & \benchhead{MVBench}
        & \benchhead{Video-MME}
        & \benchhead{LongVideoBench}
        & \benchhead{Avg.} \\
        \midrule
        & & & Acc.
        & 88.4 & 88.7 & 85.1 & 72.4 & 52.3 & 35.0 & 54.8 & 80.5 & 71.8 & 75.4 & 69.1 & 70.3 \\
        \multirow{-2}{*}{Qwen3.5}
        & \multirow{-2}{*}{\xmark} & \multirow{-2}{*}{\cmark} & Tokens
        & 1123 & 1474 & 1583 & 2955 & 3892 & 5709 & 4727 & 2034 & 1960 & 2605 & 3132 & 3050 \\
        \midrule
        & & & Acc.
        & 89.5 & 88.3 & 84.5 & 73.1 & 48.2 & 36.4 & 63.2 & 77.1 & 72.1 & 74.9 & 70.1 & 70.7 \\
        \multirow{-2}{*}{\modelname{}}
        & \multirow{-2}{*}{\xmark} & \multirow{-2}{*}{\cmark} & Tokens
        & 375 & 730 & 725 & 1446 & 1750 & 2364 & 1370 & 473 & 831 & 1372 & 1315 & 1191 \\
        \addlinespace[3pt]
        & & & Acc.
        & 88.5 & 87.8 & 83.5 & 74.2 & 50.7 & 37.7 & 63.5 & 77.4 & 70.4 & 74.4 & 69.2 & 70.7 \\
        \multirow{-2}{*}{\modelname{}}
        & \multirow{-2}{*}{\cmark} & \multirow{-2}{*}{\cmark} & Tokens
        & 415 & 797 & 749 & 1527 & 1714 & 2272 & 1204 & 512 & 857 & 1482 & 1356 & 1218 \\
        \addlinespace[3pt]
        \rowcolor{cyan!10}& & & Acc.
        & 89.5 & 89.4 & 85.1 & 75.2 & 54.2 & 39.8 & 71.0 & 81.5 & 72.4 & 76.9 & 70.5 & 73.2 \\
        \rowcolor{cyan!10}\multirow{-2}{*}{\modelname{} (default)}
        & \multirow{-2}{*}{\amark} & \multirow{-2}{*}{\amark} & Tokens
        & 330 & 766 & 740 & 653 & 848 & 76 & 37 & 429 & 542 & 1235 & 613 & 516 \\
        \bottomrule
    \end{tabular}
    \end{adjustbox}
\end{table}

\section{Qualitative Analysis}
\label{app:qualitative}

\subsection{Visual Reflection Visualization}

\begin{figure}[!htbp]
    \centering
    \includegraphics[width=\textwidth]{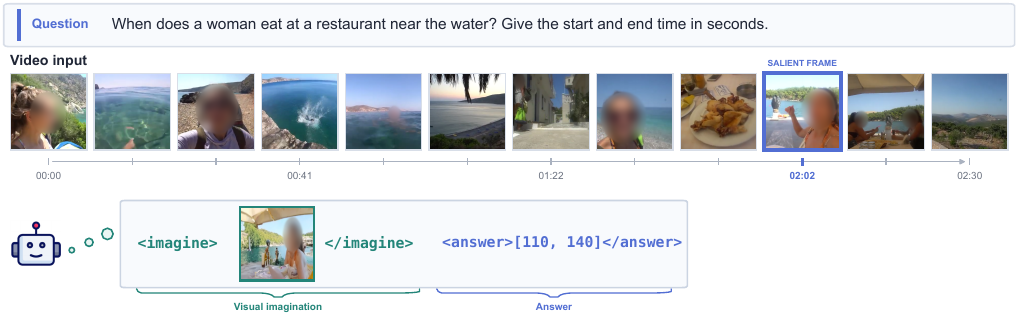}
    \caption{\textbf{Visual reflection for temporal grounding.} The rendered image
    is used only for visualization. The answer is conditioned directly on the
    continuous visual queries at inference.}
    \label{fig:video-grounding-qualitative}
\end{figure}

Figures~\ref{fig:video-qualitative} and~\ref{fig:video-grounding-qualitative}
visualize the evidence retained by the continuous visual queries for video
understanding and temporal grounding. The rendered images expose the learned visual
state for inspection, while inference uses the continuous queries directly.

\subsection{Complementary Strengths of Visual and Textual Reasoning}
\label{app:route-preferences}

\paragraph{Task-level comparison.}
We measure route preference as the accuracy difference between the forced visual and
textual routes of the same checkpoint. Figure~\ref{fig:route-preferences} shows a
structured continuum rather than a uniform advantage for either modality. The seven
largest visual margins all involve metric geometry or viewpoint transformation,
spanning room size, multi-view reasoning, relative direction, positional relations,
object distance, and object size. Art style and visual similarity show smaller but
consistent visual advantages. The strongest textual margins occur in scene
sequencing, text-referred tracking, high-resolution cross-instance reasoning,
moving direction, and spatial-relation tasks. Action localization, trajectory
reasoning, egocentric navigation, counting, functional correspondence, state
estimation, and action counting also favor textual reasoning more moderately.
Overall, visual imagination is most useful when the answer depends on reconstructing
continuous geometry, whereas textual reasoning better preserves discrete entities,
counts, and ordered state changes.

\begin{figure}[!htbp]
    \centering
    \includegraphics[width=\textwidth]{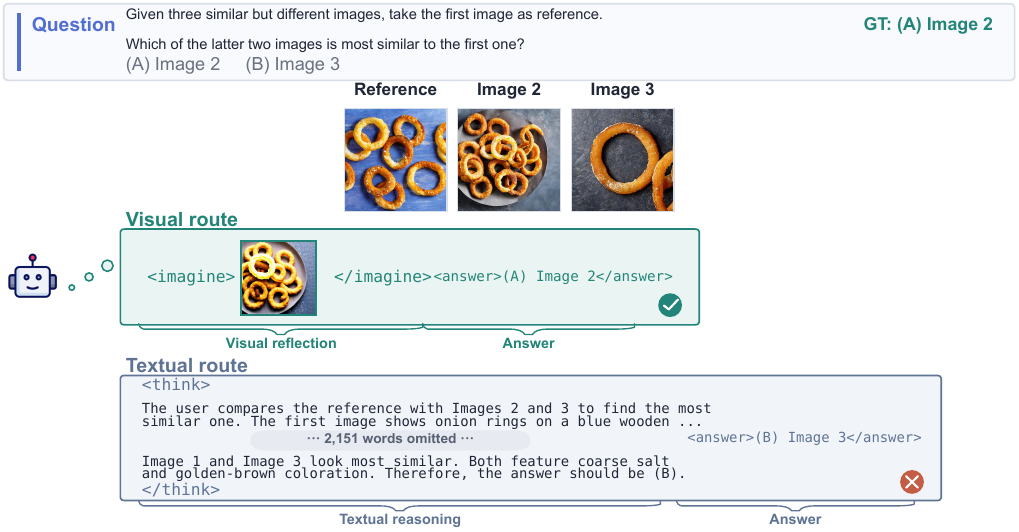}
    \caption{\textbf{Comparison between visual imagination and textual reasoning.}
    On BLINK Visual Similarity, visual imagination preserves the fine-grained shape
    evidence and answers correctly, while the long textual trace reaches the wrong
    choice.}
    \label{fig:route-visual}
\end{figure}

\begin{figure}[!htbp]
    \centering
    \includegraphics[width=\textwidth]{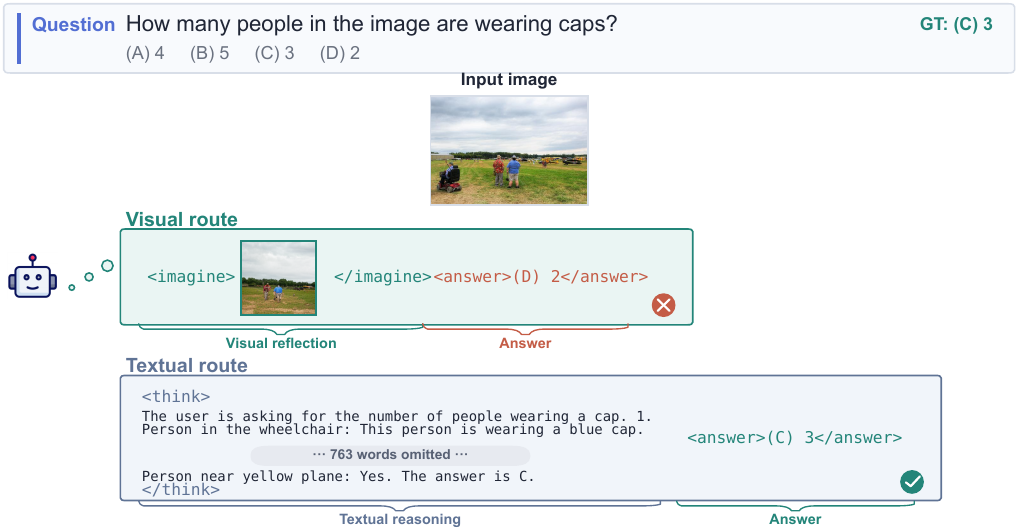}
    \caption{\textbf{Comparison between visual imagination and textual reasoning.}
    On BLINK Counting, textual reasoning returns the correct count, whereas visual
    imagination changes count-relevant details and produces an incorrect answer.}
    \label{fig:route-textual}
\end{figure}

\paragraph{Qualitative examples.}
Figures~\ref{fig:route-visual2}, \ref{fig:route-visual},
and~\ref{fig:route-textual} illustrate this complementarity. The camera-motion
example in the main paper shows that a long textual trace can still fail on a
viewpoint-sensitive spatial task. The additional visual-similarity example shows
that visual imagination can preserve fine-grained appearance, while textual
reasoning better preserves the discrete count in the counting example.

\section{Comparison With Prior Latent Visual Reasoning}
\label{app:latent-comparison}

\begin{figure}[!htbp]
    \centering
    \includegraphics[width=\textwidth]{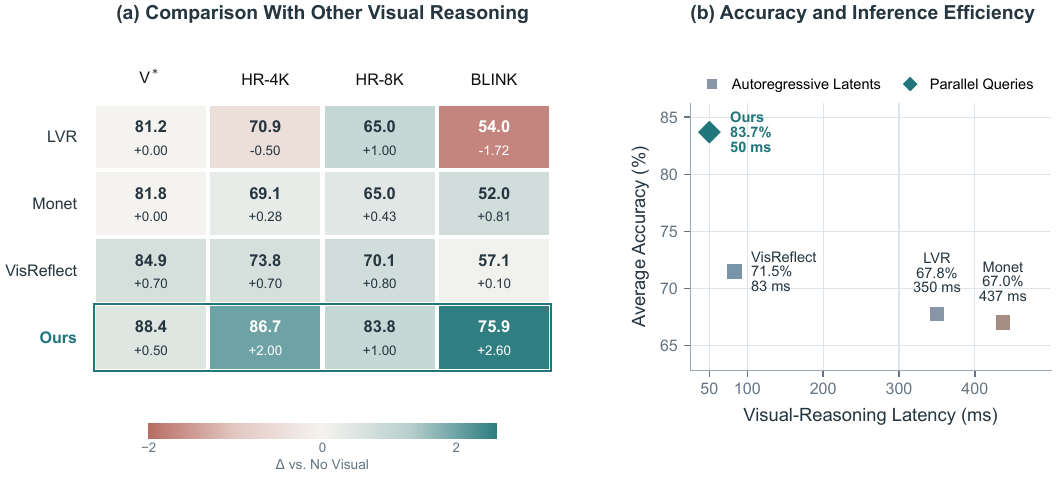}
    \caption{\textbf{Generative visual imagination improves both accuracy and
    efficiency.} (a) Accuracy on four high-resolution benchmarks; numbers
    indicate gains from enabling visual reasoning. (b) Average accuracy over the four
    benchmarks versus latency.}
    \label{fig:latent-visual-comparison}
\end{figure}

Prior latent-visual methods differ in both how their states are computed and how
those states are supervised. Figure~\ref{fig:latent-visual-comparison}(a) reports a
paired with-versus-without-visual comparison within each method. For \modelname,
these settings correspond to the Direct and fixed Visual routes in
Table~\ref{tab:ablation_mode}. Enabling our visual state improves all four shared
benchmarks, by 1.53 points on average. The corresponding mean changes are $-0.31$,
$+0.38$, and $+0.58$ for Latent Visual Reasoning (LVR), Monet, and VisReflect. For
\modelname, the largest gains occur on HR-Bench-4K (+2.0) and BLINK (+2.6),
matching the high-resolution setting for which the state is designed.

Figure~\ref{fig:latent-visual-comparison}(b) places these results against the
reported cost of constructing the visual state. \modelname reaches 83.7 average
accuracy at 50 ms, compared with 67.0 to 71.5 at 83 to 437 ms for the
autoregressive methods. The latency pattern is consistent with
the expected advantage of parallel queries over autoregressive latent generation.

\end{document}

%% file: math_commands.tex
\usepackage{amsmath,amsfonts,bm}

\def\eqref#1{equation~\ref{#1}}

\def\1{\bm{1}}

\def\vz{{\bm{z}}}

\def\mH{{\bm{H}}}

\def\mQ{{\bm{Q}}}

\DeclareMathAlphabet{\mathsfit}{\encodingdefault}{\sfdefault}{m}{sl}
\SetMathAlphabet{\mathsfit}{bold}{\encodingdefault}{\sfdefault}{bx}{n}

\def\gV{{\mathcal{V}}}



%% file: paper.bib
@article{wei2022chain,
  title={Chain-of-thought prompting elicits reasoning in large language models},
  author={Wei, Jason and Wang, Xuezhi and Schuurmans, Dale and Bosma, Maarten and Ichter, Brian and Xia, Fei and Chi, Ed and Le, Quoc V and Zhou, Denny},
  journal={Advances in Neural Information Processing Systems},
  volume={35},
  pages={24824--24837},
  year={2022}
}

@inproceedings{muennighoff2025s1,
  title={s1: Simple test-time scaling},
  author={Muennighoff, Niklas and Yang, Zitong and Shi, Weijia and Li, Xiang Lisa and Fei-Fei, Li and Hajishirzi, Hannaneh and Zettlemoyer, Luke and Liang, Percy and Cand{\`e}s, Emmanuel and Hashimoto, Tatsunori},
  booktitle={Proceedings of the 2025 Conference on Empirical Methods in Natural Language Processing},
  pages={20275--20321},
  year={2025}
}

@article{zhang2023multimodal,
  title={Multimodal chain-of-thought reasoning in language models},
  author={Zhang, Zhuosheng and Zhang, Aston and Li, Mu and Zhao, Hai and Karypis, George and Smola, Alex},
  journal={Transactions on Machine Learning Research},
  year={2023}
}

@article{shao2024visual,
  title={Visual CoT: Advancing multi-modal language models with a comprehensive dataset and benchmark for chain-of-thought reasoning},
  author={Shao, Hao and Qian, Shengju and Xiao, Han and Song, Guanglu and Zong, Zhuofan and Wang, Letian and Liu, Yu and Li, Hongsheng},
  journal={Advances in Neural Information Processing Systems},
  volume={37},
  pages={8612--8642},
  year={2024}
}

@article{hu2024visual,
  title={Visual sketchpad: Sketching as a visual chain of thought for multimodal language models},
  author={Hu, Yushi and Shi, Weijia and Fu, Xingyu and Roth, Dan and Ostendorf, Mari and Zettlemoyer, Luke and Smith, Noah A and Krishna, Ranjay},
  journal={Advances in Neural Information Processing Systems},
  volume={37},
  pages={139348--139379},
  year={2024}
}

@inproceedings{gupta2023visual,
  title={Visual programming: Compositional visual reasoning without training},
  author={Gupta, Tanmay and Kembhavi, Aniruddha},
  booktitle={Proceedings of the IEEE/CVF Conference on Computer Vision and Pattern Recognition},
  pages={14953--14962},
  year={2023}
}

@inproceedings{suris2023vipergpt,
  title={ViperGPT: Visual inference via Python execution for reasoning},
  author={Sur{\'\i}s, D{\'\i}dac and Menon, Sachit and Vondrick, Carl},
  booktitle={Proceedings of the IEEE/CVF International Conference on Computer Vision},
  pages={11888--11898},
  year={2023}
}

@article{yang2023set,
  title={Set-of-Mark prompting unleashes extraordinary visual grounding in GPT-4V},
  author={Yang, Jianwei and Zhang, Hao and Li, Feng and Zou, Xueyan and Li, Chunyuan and Gao, Jianfeng},
  journal={arXiv preprint arXiv:2310.11441},
  year={2023}
}

@inproceedings{li2025imagine,
  title={Imagine while reasoning in space: Multimodal visualization-of-thought},
  author={Li, Chengzu and Wu, Wenshan and Zhang, Huanyu and Xia, Yan and Mao, Shaoguang and Dong, Li and Vuli{\'c}, Ivan and Wei, Furu},
  booktitle={Proceedings of the 42nd International Conference on Machine Learning},
  volume={267},
  pages={36340--36364},
  publisher={PMLR},
  year={2025}
}

@article{hao2024training,
  title={Training large language models to reason in a continuous latent space},
  author={Hao, Shibo and Sukhbaatar, Sainbayar and Su, DiJia and Li, Xian and Hu, Zhiting and Weston, Jason and Tian, Yuandong},
  journal={arXiv preprint arXiv:2412.06769},
  year={2024}
}

@inproceedings{goyal2024think,
  title={Think before you speak: Training language models with pause tokens},
  author={Goyal, Sachin and Ji, Ziwei and Rawat, Ankit Singh and Menon, Aditya Krishna and Kumar, Sanjiv and Nagarajan, Vaishnavh},
  booktitle={International Conference on Learning Representations},
  volume={2024},
  pages={27896--27923},
  year={2024}
}

@article{pan2025transfer,
  title={Transfer between modalities with MetaQueries},
  author={Pan, Xichen and Shukla, Satya Narayan and Singh, Aashu and others},
  journal={arXiv preprint arXiv:2504.06256},
  year={2025}
}

@inproceedings{zhou2025transfusion,
  title={Transfusion: Predict the next token and diffuse images with one multi-modal model},
  author={Zhou, Chunting and Yu, Lili and Babu, Arun and others},
  booktitle={International Conference on Learning Representations},
  volume={2025},
  pages={6446--6469},
  year={2025}
}

@inproceedings{xie2025showo,
  title={Show-o: One single transformer to unify multimodal understanding and generation},
  author={Xie, Jinheng and Mao, Weijia and Bai, Zechen and others},
  booktitle={International Conference on Learning Representations},
  volume={2025},
  pages={28240--28264},
  year={2025}
}

@inproceedings{dong2024dreamllm,
  title={DreamLLM: Synergistic multimodal comprehension and creation},
  author={Dong, Runpei and Han, Chunrui and Peng, Yuang and Qi, Zekun and Ge, Zheng and Yang, Jinrong and Zhao, Liang and Sun, Jianjian and Zhou, Hongyu and Wei, Haoran and Kong, Xiangwen and others},
  booktitle={International Conference on Learning Representations},
  volume={2024},
  pages={6666--6702},
  year={2024}
}

@article{liu2023visual,
  title={Visual instruction tuning},
  author={Liu, Haotian and Li, Chunyuan and Wu, Qingyang and Lee, Yong Jae},
  journal={Advances in Neural Information Processing Systems},
  volume={36},
  pages={34892--34916},
  year={2023}
}

@article{bai2025qwen25vl,
  title={Qwen2.5-VL technical report},
  author={Bai, Shuai and Chen, Keqin and Liu, Xuejing and others},
  journal={arXiv preprint arXiv:2502.13923},
  year={2025}
}

@article{li2024llavaonevision,
  title={LLaVA-OneVision: Easy visual task transfer},
  author={Li, Bo and Zhang, Yuanhan and Guo, Dong and Zhang, Renrui and Li, Feng and Zhang, Hao and Zhang, Kaichen and Zhang, Peiyuan and Li, Yanwei and Liu, Ziwei and Li, Chunyuan},
  journal={arXiv preprint arXiv:2408.03326},
  year={2024}
}

@article{tong2024cambrian,
  title={Cambrian-1: A fully open, vision-centric exploration of multimodal LLMs},
  author={Tong, Shengbang and Brown, Ellis and Wu, Penghao and others},
  journal={Advances in Neural Information Processing Systems},
  volume={37},
  pages={87310--87356},
  year={2024}
}

@article{shao2024deepseekmath,
  title={DeepSeekMath: Pushing the limits of mathematical reasoning in open language models},
  author={Shao, Zhihong and Wang, Peiyi and Zhu, Qihao and others},
  journal={arXiv preprint arXiv:2402.03300},
  year={2024}
}

@inproceedings{chen2025not,
  title={Do NOT think that much for 2+3=? On the overthinking of long reasoning models},
  author={Chen, Xingyu and Xu, Jiahao and Liang, Tian and others},
  booktitle={Proceedings of the 42nd International Conference on Machine Learning},
  volume={267},
  pages={9487--9499},
  publisher={PMLR},
  year={2025}
}

@article{sui2025stop,
  title={Stop overthinking: A survey on efficient reasoning for large language models},
  author={Sui, Yang and Chuang, Yu-Neng and Wang, Guanchu and others},
  journal={Transactions on Machine Learning Research},
  year={2025}
}

@article{fang2025thinkless,
  title={Thinkless: LLM learns when to think},
  author={Fang, Gongfan and Ma, Xinyin and Wang, Xinchao},
  journal={Advances in Neural Information Processing Systems},
  volume={38},
  pages={151268--151295},
  year={2025}
}

@misc{qwen2026qwen359b,
  title={{Qwen3.5-9B}},
  author={{Qwen Team}},
  year={2026},
  note={Hugging Face model card},
  howpublished={\url{https://huggingface.co/Qwen/Qwen3.5-9B}}
}

@misc{qwen2026qwen352b,
  title={{Qwen3.5-2B}},
  author={{Qwen Team}},
  year={2026},
  note={Hugging Face model card},
  howpublished={\url{https://huggingface.co/Qwen/Qwen3.5-2B}}
}

@misc{qwen2026qwen3527b,
  title={{Qwen3.5-27B}},
  author={{Qwen Team}},
  year={2026},
  note={Hugging Face model card},
  howpublished={\url{https://huggingface.co/Qwen/Qwen3.5-27B}}
}

@inproceedings{fu2024blink,
  title={BLINK: Multimodal large language models can see but not perceive},
  author={Fu, Xingyu and Hu, Yushi and Li, Bangzheng and others},
  booktitle={European Conference on Computer Vision},
  pages={148--166},
  year={2024},
  organization={Springer}
}

@inproceedings{fu2025videomme,
  title={Video-MME: The first-ever comprehensive evaluation benchmark of multi-modal LLMs in video analysis},
  author={Fu, Chaoyou and Dai, Yuhan and Luo, Yongdong and others},
  booktitle={Proceedings of the Computer Vision and Pattern Recognition Conference (CVPR)},
  pages={24108--24118},
  year={2025}
}

@inproceedings{li2024mvbench,
  title={MVBench: A comprehensive multi-modal video understanding benchmark},
  author={Li, Kunchang and Wang, Yali and He, Yinan and others},
  booktitle={Proceedings of the IEEE/CVF Conference on Computer Vision and Pattern Recognition},
  pages={22195--22206},
  year={2024}
}

@article{wu2024longvideobench,
  title={LongVideoBench: A benchmark for long-context interleaved video-language understanding},
  author={Wu, Haoning and Li, Dongxu and Chen, Bei and Li, Junnan},
  journal={Advances in Neural Information Processing Systems},
  volume={37},
  pages={28828--28857},
  year={2024}
}

@inproceedings{yang2025thinking,
  title={Thinking in space: How multimodal large language models see, remember, and recall spaces},
  author={Yang, Jihan and Yang, Shusheng and Gupta, Anjali W and Han, Rilyn and Fei-Fei, Li and Xie, Saining},
  booktitle={Proceedings of the IEEE/CVF Conference on Computer Vision and Pattern Recognition},
  pages={10632--10643},
  year={2025}
}

@inproceedings{gao2017tall,
  title={TALL: Temporal activity localization via language query},
  author={Gao, Jiyang and Sun, Chen and Yang, Zhenheng and Nevatia, Ram},
  booktitle={Proceedings of the IEEE International Conference on Computer Vision},
  pages={5267--5275},
  year={2017}
}

@article{lei2021detecting,
  title={Detecting moments and highlights in videos via natural language queries},
  author={Lei, Jie and Berg, Tamara L and Bansal, Mohit},
  journal={Advances in Neural Information Processing Systems},
  volume={34},
  pages={11846--11858},
  year={2021}
}

@inproceedings{xiao2024nextgqa,
  title={Can I trust your answer? Visually grounded video question answering},
  author={Xiao, Junbin and Yao, Angela and Li, Yicong and Chua, Tat-Seng},
  booktitle={Proceedings of the IEEE/CVF Conference on Computer Vision and Pattern Recognition},
  pages={13204--13214},
  year={2024}
}

@article{duan2026joyai,
  title={JoyAI-Image: Awaking Spatial Intelligence in Unified Multimodal Understanding and Generation},
  author={Song, Lin and Li, Wenbo and Ma, Guoqing and Tang, Wei and Wang, Bo and Zhang, Yuan and others},
  journal={arXiv preprint arXiv:2605.04128},
  year={2026}
}

@article{tong2025openo3,
  title={Open-o3-Video: Grounded Video Reasoning with Explicit Spatio-Temporal Evidence},
  author={Meng, Jiahao and Li, Xiangtai and Wang, Haochen and Tan, Yue and Zhang, Tao and Kong, Lingdong and Tong, Yunhai and others},
  journal={arXiv preprint arXiv:2510.20579},
  year={2025}
}

@article{videochat3team2026,
  title={VideoChat3: Fully Open Video MLLM for Efficient and Generalist Video Understanding},
  author={Li, Xinhao and Zhu, Yuhan and Zeng, Xiangyu and Dong, Yuhao and Wu, Haoning and Zhang, Zhiqiu and others},
  journal={arXiv preprint arXiv:2607.14935},
  year={2026}
}

@article{erqa2025,
  title={Gemini Robotics: Bringing AI into the Physical World},
  author={{Gemini Robotics Team} and Abeyruwan, Saminda and Ainslie, Joshua and others},
  journal={arXiv preprint arXiv:2503.20020},
  year={2025}
}

@inproceedings{yang2026mmsi,
  title={Mmsi-bench: A benchmark for multi-image spatial intelligence},
  author={Yang, Sihan and Xu, Runsen and Xie, Yiman and Yang, Sizhe and Li, Mo and Lin, Jingli and Zhu, Chenming and Chen, Xiaochen and Duan, Haodong and Yue, Xiangyu and others},
  booktitle={International Conference on Learning Representations},
  volume={2026},
  pages={157051--157088},
  year={2026}
}

@misc{xai2024realworldqa,
  title={Grok-1.5 Vision Preview},
  author={{xAI}},
  year={2024},
  note={Introduces the RealWorldQA benchmark},
  howpublished={\url{https://x.ai/news/grok-1.5v}}
}

@inproceedings{zhou2025mlvu,
  title={MLVU: Benchmarking Multi-task Long Video Understanding},
  author={Zhou, Junjie and Shu, Yan and Zhao, Bo and Wu, Boya and Liang, Zhengyang and Xiao, Shitao and Qin, Minghao and Yang, Xi and Xiong, Yongping and Zhang, Bo and Huang, Tiejun and Liu, Zheng},
  booktitle={Proceedings of the IEEE/CVF Conference on Computer Vision and Pattern Recognition},
  pages={13691--13701},
  year={2025}
}

@inproceedings{li2026latent,
  title={Latent visual reasoning},
  author={Li, Bangzheng and Sun, Ximeng and Liu, Jiang and Wang, Ze and Wu, Jialian and Yu, Xiaodong and Barsoum, Emad and Chen, Muhao and Liu, Zicheng},
  booktitle={International Conference on Learning Representations},
  volume={2026},
  pages={148076--148090},
  year={2026}
}

@inproceedings{yang2026machine,
  title={Machine mental imagery: Empower multimodal reasoning with latent visual tokens},
  author={Yang, Zeyuan and Yu, Xueyang and Chen, Delin and Shen, Maohao and Gan, Chuang},
  booktitle={Proceedings of the IEEE/CVF Conference on Computer Vision and Pattern Recognition},
  pages={33510--33520},
  year={2026}
}

@article{pham2025multimodal,
  title={Multimodal chain of continuous thought for latent-space reasoning in vision-language models},
  author={Pham, Tan-Hanh and Ngo, Chris},
  journal={arXiv preprint arXiv:2508.12587},
  year={2025}
}

@inproceedings{wang2026monet,
  title={Monet: Reasoning in latent visual space beyond image and language},
  author={Wang, Qixun and Shi, Yang and Wang, Yifei and Zhang, Yuanxing and Wan, Pengfei and Gai, Kun and Ying, Xianghua and Wang, Yisen},
  booktitle={Proceedings of the IEEE/CVF Conference on Computer Vision and Pattern Recognition},
  pages={12030--12040},
  year={2026}
}

@inproceedings{liu2026videoauto,
  title={VideoAuto-R1: Video Auto Reasoning via Thinking Once, Answering Twice},
  author={Liu, Shuming and Zhuge, Mingchen and Zhao, Changsheng and Chen, Jun and Wu, Lemeng and Liu, Zechun and Zhu, Chenchen and Cai, Zhipeng and Zhou, Chong and others},
  booktitle={Proceedings of the IEEE/CVF Conference on Computer Vision and Pattern Recognition},
  pages={32829--32839},
  year={2026}
}

@inproceedings{shen2026visreflect,
  title={VisReflect: Latent Visual Reflection for Fine-Grained Perception in Long Visual Context},
  author={Shen, Xiaoqian and Elhoseiny, Mohamed},
  booktitle={European Conference on Computer Vision},
  pages={632--649},
  year={2026},
  organization={Springer}
}

@inproceedings{kang2025see,
  title={See what you are told: Visual attention sink in large multimodal models},
  author={Kang, Seil and Kim, Jinyeong and Kim, Junhyeok and Hwang, Seong Jae},
  booktitle={International Conference on Learning Representations},
  volume={2025},
  pages={87676--87703},
  year={2025}
}

@inproceedings{wu2024vstar,
  title={V*: Guided visual search as a core mechanism in multimodal LLMs},
  author={Wu, Penghao and Xie, Saining},
  booktitle={Proceedings of the IEEE/CVF Conference on Computer Vision and Pattern Recognition},
  pages={13084--13094},
  year={2024}
}

@article{wang2025divide,
  title={Divide, conquer and combine: A training-free framework for high-resolution image perception in multimodal large language models},
  author={Wang, Wenbin and Ding, Liang and Zeng, Minyan and Zhou, Xiabin and Shen, Li and Luo, Yong and Yu, Wei and Tao, Dacheng},
  journal={Proceedings of the AAAI Conference on Artificial Intelligence},
  volume={39},
  number={8},
  pages={7907--7915},
  year={2025}
}

@inproceedings{zheng2026deepeyes,
  title={DeepEyes: Incentivizing ``thinking with images'' via reinforcement learning},
  author={Zheng, Ziwei and Yang, Minghao and Hong, Jack and Zhao, Chenxiao and Xu, Guohai and Yang, Le and Shen, Chao and Yu, Xing},
  booktitle={International Conference on Learning Representations},
  volume={2026},
  pages={126775--126798},
  year={2026}
}

@article{su2026pixel,
  title={Pixel reasoner: Incentivizing pixel space reasoning via curiosity-driven reinforcement learning},
  author={Su, Alex and Wang, Haozhe and Ren, Weiming and Lin, Fangzhen and Chen, Wenhu},
  journal={Advances in Neural Information Processing Systems},
  volume={38},
  pages={8222--8251},
  year={2026}
}

@article{feng2025video,
  title={Video-R1: Reinforcing video reasoning in MLLMs},
  author={Feng, Kaituo and Gong, Kaixiong and Li, Bohao and Guo, Zonghao and Wang, Yibing and Peng, Tianshuo and Wu, Junfei and Zhang, Xiaoying and Wang, Benyou and Yue, Xiangyu},
  journal={Advances in Neural Information Processing Systems},
  volume={38},
  pages={99114--99137},
  year={2025}
}

@article{yan2026videochat,
  title={Videochat-r1. 5: Visual test-time scaling to reinforce multimodal reasoning by iterative perception},
  author={Yan, Ziang and He, Yinan and Li, Xinhao and Yue, Zhengrong and Zeng, Xiangyu and Wang, Yali and Qiao, Yu and Wang, Limin and Wang, Yi},
  journal={Advances in Neural Information Processing Systems},
  volume={38},
  pages={119152--119184},
  year={2026}
}

@article{wang2025internvl35,
  title={InternVL3.5: Advancing open-source multimodal models in versatility, reasoning, and efficiency},
  author={Wang, Weiyun and Gao, Zhangwei and Gu, Lixin and Pu, Hengjun and Cui, Long and Wei, Xingguang and Liu, Zhaoyang and Jing, Linglin and Ye, Shenglong and Shao, Jie and others},
  journal={arXiv preprint arXiv:2508.18265},
  year={2025}
}

@article{liu2025understanding,
  title={Understanding r1-zero-like training: A critical perspective},
  author={Liu, Zichen and Chen, Changyu and Li, Wenjun and Qi, Penghui and Pang, Tianyu and Du, Chao and Lee, Wee Sun and Lin, Min},
  journal={arXiv preprint arXiv:2503.20783},
  year={2025}
}

@inproceedings{zhang2025adaptthink,
  title={AdaptThink: Reasoning Models Can Learn When to Think},
  author={Zhang, Jiajie and Lin, Nianyi and Hou, Lei and Feng, Ling and Li, Juanzi},
  booktitle={Proceedings of the 2025 Conference on Empirical Methods in Natural Language Processing},
  pages={3716--3730},
  publisher={Association for Computational Linguistics},
  year={2025}
}

@inproceedings{xie2025sana,
  title={{SANA}: Efficient high-resolution text-to-image synthesis with linear diffusion transformers},
  author={Xie, Enze and Chen, Junsong and Chen, Junyu and Cai, Han and Tang, Haotian and Lin, Yujun and Zhang, Zhekai and Li, Muyang and Zhu, Ligeng and Lu, Yao and Han, Song},
  booktitle={International Conference on Learning Representations},
  volume={2025},
  pages={20383--20407},
  year={2025}
}

@article{shen2025zoom,
  title={Zoom-zero: Reinforced coarse-to-fine video understanding via temporal zoom-in},
  author={Shen, Xiaoqian and Chen, Min-Hung and Wang, Yu-Chiang Frank and Elhoseiny, Mohamed and Hachiuma, Ryo},
  journal={arXiv preprint arXiv:2512.14273},
  year={2025}
}

@article{openai2025gpt5,
  title={Openai gpt-5 system card},
  author={Singh, Aaditya and Fry, Adam and Perelman, Adam and Tart, Adam and Ganesh, Adi and El-Kishky, Ahmed and McLaughlin, Aidan and Low, Aiden and Ostrow, AJ and Ananthram, Akhila and others},
  journal={arXiv preprint arXiv:2601.03267},
  year={2025}
}

@misc{openai2025gpt5mini,
  title={GPT-5 mini Model},
  author={{OpenAI}},
  year={2025},
  howpublished={\url{https://developers.openai.com/api/docs/models/gpt-5-mini}}
}

@article{comanici2025gemini25,
  title={Gemini 2.5: Pushing the Frontier with Advanced Reasoning, Multimodality, Long Context, and Next Generation Agentic Capabilities},
  author={Comanici, Gheorghe and Bieber, Eric and Schaekermann, Mike and Pasupat, Ice and Sachdeva, Noveen and Dhillon, Inderjit and others},
  journal={arXiv preprint arXiv:2507.06261},
  year={2025}
}

@article{yang2025cambrians,
  title={Cambrian-S: Towards Spatial Supersensing in Video},
  author={Yang, Shusheng and Yang, Jihan and Huang, Pinzhi and Brown, Ellis and Yang, Zihao and Yu, Yue and others},
  journal={arXiv preprint arXiv:2511.04670},
  year={2025}
}

@article{an2026llavaonevision2,
  title={LLaVA-OneVision-2: Towards Next-Generation Perceptual Intelligence},
  author={An, Xiang and Xie, Yin and Tang, Feilong and Yan, Yunyao and Tan, Huajie and Zhu, Didi and others},
  journal={arXiv preprint arXiv:2605.25979},
  year={2026}
}

@article{gemmateam2026gemma4,
  title={Gemma 4 Technical Report},
  author={{Gemma Team} and El Abd, Sherif and Aggarwal, Vaibhav and Algayres, Robin and others},
  journal={arXiv preprint arXiv:2607.02770},
  year={2026}
}

@misc{openbmb2026minicpmv46,
  title={{MiniCPM-V} 4.6},
  author={{OpenBMB}},
  year={2026},
  note={Hugging Face model card},
  howpublished={\url{https://huggingface.co/openbmb/MiniCPM-V-4.6}}
}

@article{wang2026timer1,
  title={Time-R1: Post-Training Large Vision Language Model for Temporal Video Grounding},
  author={Wang, Ye and Wang, Ziheng and Xu, Boshen and Du, Yang and Lin, Kejun and Xiao, Zihan and others},
  journal={Advances in Neural Information Processing Systems},
  volume={38},
  pages={83330--83364},
  year={2026}
}

@inproceedings{zhang2026vital,
  title={Thinking With Videos: Multimodal Tool-Augmented Reinforcement Learning for Long Video Reasoning},
  author={Zhang, Haoji and Gu, Xin and Li, Jiawen and Ma, Chixiang and Bai, Sule and Zhang, Chubin and others},
  booktitle={Proceedings of the IEEE/CVF Conference on Computer Vision and Pattern Recognition},
  pages={32903--32914},
  year={2026}
}

@article{chen2025blip3o,
  title={BLIP3-o: A Family of Fully Open Unified Multimodal Models, Architecture, Training and Dataset},
  author={Chen, Jiuhai and Xu, Zhiyang and Pan, Xichen and Hu, Yushi and Qin, Can and Goldstein, Tom and others},
  journal={arXiv preprint arXiv:2505.09568},
  year={2025}
}

@inproceedings{changpinyo2021conceptual,
  title={Conceptual 12M: Pushing Web-Scale Image-Text Pre-Training to Recognize Long-Tail Visual Concepts},
  author={Changpinyo, Soravit and Sharma, Piyush and Ding, Nan and Soricut, Radu},
  booktitle={Proceedings of the IEEE/CVF Conference on Computer Vision and Pattern Recognition},
  pages={3558--3568},
  year={2021}
}

@inproceedings{kirillov2023segment,
  title={Segment Anything},
  author={Kirillov, Alexander and Mintun, Eric and Ravi, Nikhila and Mao, Hanzi and Rolland, Chloe and Gustafson, Laura and others},
  booktitle={Proceedings of the IEEE/CVF International Conference on Computer Vision},
  pages={4015--4026},
  year={2023}
}

@inproceedings{krishna2017dense,
  title={Dense-Captioning Events in Videos},
  author={Krishna, Ranjay and Hata, Kenji and Ren, Frederic and Fei-Fei, Li and Niebles, Juan Carlos},
  booktitle={Proceedings of the IEEE International Conference on Computer Vision},
  pages={706--715},
  year={2017}
}

@inproceedings{chen2024internvl,
  title={InternVL: Scaling up Vision Foundation Models and Aligning for Generic Visual-Linguistic Tasks},
  author={Chen, Zhe and Wu, Jiannan and Wang, Wenhai and Su, Weijie and Chen, Guo and Xing, Sen and others},
  booktitle={Proceedings of the IEEE/CVF Conference on Computer Vision and Pattern Recognition},
  pages={24185--24198},
  year={2024}
}

@article{chen2024far,
  title={How far are we to {GPT-4V}? Closing the gap to commercial multimodal models with open-source suites},
  author={Chen, Zhe and Wang, Weiyun and Tian, Hao and Ye, Shenglong and Gao, Zhangwei and Cui, Erfei and Tong, Wenwen and Hu, Kongzhi and Luo, Jiapeng and Ma, Zheng and Ma, Ji and Wang, Jiaqi and Dong, Xiaoyi and Yan, Hang and Guo, Hewei and He, Conghui and Shi, Botian and Jin, Zhenjiang and Xu, Chao and Wang, Bin and Wei, Xingjian and Li, Wei and Zhang, Wenjian and Zhang, Bo and Cai, Pinlong and Wen, Licheng and Yan, Xiangchao and Dou, Min and Lu, Lewei and Zhu, Xizhou and Lu, Tong and Lin, Dahua and Qiao, Yu and Dai, Jifeng and Wang, Wenhai},
  journal={Science China Information Sciences},
  volume={67},
  number={12},
  pages={220101},
  year={2024},
  doi={10.1007/s11432-024-4231-5},
  url={https://doi.org/10.1007/s11432-024-4231-5}
}

@inproceedings{zhu2024minigpt,
  title={MiniGPT-4: Enhancing vision-language understanding with advanced large language models},
  author={Zhu, Deyao and Chen, Jun and Shen, Xiaoqian and Li, Xiang and Elhoseiny, Mohamed},
  booktitle={International Conference on Learning Representations},
  volume={2024},
  pages={18378--18394},
  year={2024}
}

@article{jeon2026vision,
  title={Vision-aligned latent reasoning for multi-modal large language model},
  author={Jeon, Byungwoo and Jeong, Yoonwoo and Lee, Hyunseok and Cho, Minsu and Shin, Jinwoo},
  journal={arXiv preprint arXiv:2602.04476},
  year={2026}
}

@inproceedings{loshchilov2019decoupled,
  title={Decoupled weight decay regularization},
  author={Loshchilov, Ilya and Hutter, Frank},
  booktitle={International Conference on Learning Representations},
  year={2019},
  url={https://openreview.net/forum?id=Bkg6RiCqY7}
}

@article{wang2004image,
  title={Image Quality Assessment: From Error Visibility to Structural Similarity},
  author={Wang, Zhou and Bovik, Alan C. and Sheikh, Hamid R. and Simoncelli, Eero P.},
  journal={IEEE Transactions on Image Processing},
  volume={13},
  number={4},
  pages={600--612},
  year={2004},
  doi={10.1109/TIP.2003.819861}
}
